\documentclass{article}

\usepackage{arxiv}

\usepackage[utf8]{inputenc} % allow utf-8 input
\usepackage[T1]{fontenc}    % use 8-bit T1 fonts
\usepackage{hyperref}       % hyperlinks
\usepackage{url}            % simple URL typesetting
\usepackage{booktabs}       % professional-quality tables
\usepackage{amsfonts}       % blackboard math symbols
\usepackage{nicefrac}       % compact symbols for 1/2, etc.
\usepackage{microtype}      % microtypography
\usepackage{lipsum}		% Can be removed after putting your text content
\usepackage{graphicx}
\usepackage{natbib}
\usepackage{doi}

\usepackage[font=bf]{caption} %make captions bold
\usepackage{float} %for allowing figures to float

\usepackage{color}
\usepackage{amsmath}
\usepackage{amsthm}
\usepackage{amsfonts}
\usepackage{natbib}
\usepackage[noend]{algpseudocode}
\usepackage{tikz}
\usepackage{xcolor}
\usepackage{color, colortbl}
\usepackage{amssymb}
\usepackage[nodisplayskipstretch]{setspace}
\usepackage{subfig}
\usepackage{xcolor}
\usepackage[linesnumbered,ruled,vlined]{algorithm2e}

\SetCommentSty{mycommfont}
\usepackage[noend]{algpseudocode}
\SetKwInput{KwInput}{Input}                % Set the Input
\SetKwInput{KwOutput}{Output}              % set the Output
\SetKwInput{KwParameter}{Parameter} 

\usetikzlibrary{shadows}
\tikzset{
diagonal fill/.style 2 args={fill=#2, path picture={
\fill[#1, sharp corners] (path picture bounding box.south west) -|
                         (path picture bounding box.north east) -- cycle;}},
reversed diagonal fill/.style 2 args={fill=#2, path picture={
\fill[#1, sharp corners] (path picture bounding box.north west) |- 
                         (path picture bounding box.south east) -- cycle;}}
}

\usetikzlibrary{bayesnet}
\usetikzlibrary{arrows}
\usetikzlibrary{calc}
\usetikzlibrary{shapes,decorations,arrows,calc,arrows.meta,fit,positioning, automata}
\tikzset{
    -Latex,auto,node distance =1 cm and 1 cm,semithick,
    state/.style ={ellipse, draw, minimum width = 0.7 cm},
    point/.style = {circle, draw, inner sep=0.04cm,fill,node contents={}},
    bidirected/.style={Latex-Latex,dashed},
    el/.style = {inner sep=2pt, align=left, sloped}
}
\algnewcommand{\LineComment}[1]{ \(\blacktriangleright\) \emph{\color{brown} #1}}

\def\eqref#1{Eq.~(\ref{#1})}
\def\figref#1{Fig.~\ref{#1}}

\newcommand{\indep}{\perp \!\!\! \perp}

\newcommand*\xbar[1]{%
  \hbox{%
    \vbox{%
      \hrule height 0.5pt % The actual bar
      \kern0.5ex%         % Distance between bar and symbol
      \hbox{%
        \kern-0.1em%      % Shortening on the left side
        \ensuremath{#1}%
        \kern-0.1em%      % Shortening on the right side
      }%
    }%
  }%
}

\def\algref#1{line~\ref{#1}}

\title{Experimental design for causal query estimation in partially observed biomolecular networks}

\author{ \href{https://orcid.org/0000-0002-6554-9083}{\includegraphics[scale=0.06]{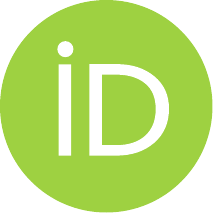}\hspace{1mm}Sara ~Mohammad-Taheri}\thanks{Corresponding author.} \\
	Khoury College of Computer Sciences\\
	Northeastern University\\
	Boston, MA 02115 \\
	\texttt{mohammadtaheri.s@northeastern.edu} \\
	\And
	\href{https://orcid.org/0000-0002-7276-9009}{\includegraphics[scale=0.06]{orcid.pdf}\hspace{1mm}Jeremy ~Zucker} \\
	Pacific Northwest National Laboratory\\
	Richland, WA 99354\\
	\texttt{jeremy.zucker@pnnl.gov} \\
	\And
	\href{https://orcid.org/0000-0000-0000-0000}{\includegraphics[scale=0.06]{orcid.pdf}\hspace{1mm}Charles Tapley ~Hoyt} \\
	Laboratory of Systems Pharmacology\\
	Harvard Medical School\\
	Boston, MA\\
	\texttt{cthoyt@gmail.com} \\
	\And
	{\includegraphics[scale=0.06]{orcid.pdf}\hspace{1mm}Karen ~Sachs} \\
	Next Generation Analytics\\
	Palo Alto, CA\\
	\texttt{sachskaren@gmail.com} \\
	\And
	\href{https://orcid.org/0000-0003-0002-4259}{\includegraphics[scale=0.06]{orcid.pdf}\hspace{1mm}Vartika ~Tewari} \\
	Khoury College of Computer Sciences\\
	Northeastern University\\
	Boston, MA 02115 \\
	\texttt{tewari.v@northeastern.edu} \\
	\And
	{\includegraphics[scale=0.06]{orcid.pdf}\hspace{1mm}Robert ~Ness} \\
	Microsoft Research\\
	Redmod, WA \\
	\texttt{robertness@gmail.com} \\
	\And
	\href{https://orcid.org/0000-0003-1728-1104}{\includegraphics[scale=0.06]{orcid.pdf}\hspace{1mm}Olga ~Vitek} \\
	\thanks{Corresponding author.} \\
	Khoury College of Computer Sciences\\
	Northeastern University\\
	Boston, MA 02115 \\
	\texttt{o.vitek@northeastern.edu} \\
}

\renewcommand{\shorttitle}{\textit{arXiv} Template}

\hypersetup{
pdftitle={A template for the arxiv style},
pdfsubject={q-bio.NC, q-bio.QM},
pdfauthor={David S.~Hippocampus, Elias D.~Striatum},
pdfkeywords={First keyword, Second keyword, More},
}

\begin{document}
\maketitle

\begin{abstract}
\textbf{Motivation:} Estimating causal queries, such as changes in protein abundance in response to a perturbation, is a fundamental task in the analysis of biomolecular pathways.  
The estimation requires experimental measurements on the pathway components. However, in practice many pathway components are left unobserved (latent) because they are either unknown, or difficult to measure.
Latent variable models (LVMs) are well-suited for such estimation. 
Unfortunately, LVM-based estimation of causal queries can be inaccurate when parameters of the latent variables are not uniquely identified, or when the number of latent variables is misspecified. This has limited the use of LVMs for causal inference in biomolecular pathways.\\
\textbf{Results:} In this manuscript, we propose a general and practical approach for LVM-based estimation of causal queries.
We prove that, despite the challenges above, LVM-based estimators of causal queries are accurate if the queries are identifiable according to Pearl's do-calculus, and describe an algorithm for its estimation.
We illustrate the breadth and the practical utility of this approach for estimating causal queries in four synthetic and two experimental case studies, where structures of biomolecular pathways challenge the existing methods for causal query estimation. \\
\textbf{Availability:}  
The code and the data documenting all the case studies are available at \url{https://github.com/srtaheri/LVMwithDoCalculus}\\
\textbf{Contact:} \href{o.vitek@northeastern.edu}{o.vitek@northeastern.edu}\\
\textbf{Supplementary information:} Supplementary Materials are available as a separate document. 
\end{abstract}

% keywords can be removed

\section{Introduction}

Biomolecular pathways are governed by complex patterns of controls such as signaling, gene regulation, and metabolic reactions. Biomolecular pathways are often represented as graphs, where nodes are signaling proteins, genes, transcripts or metabolites, and directed edges are causal regulatory relationships.
The graph-based representations are useful for simulating experimental perturbations, and answering, \emph{in silico},  causal queries of the form {\it ``When we perturb $X$, what is the effect on its descendent $Y$?"}. However, estimation of causal queries requires more than a qualitative topology of the graph. It also requires experimental measurements on the nodes of the graph, in order to quantitatively characterize the causal relationships~\cite{pearl2009causality}.

Unfortunately, no measurement modality can currently capture all the molecular components of a pathway. The incomplete data arise in at least two general, ubiquitous scenarios. The first occurs when components of a biomolecular pathway are not fully known. For example, there may be empirical evidence for the regulation of an enzyme, but the identity of the molecule or protein that regulates the enzyme may be unknown~\cite{cannon_2021}. The second scenario occurs when, due to limitations of the measurement techniques, some pathway components are unobserved. For example, antibodies for a protein may not be available. Alternatively, while RNA abundances may be characterized, levels of the corresponding  protein or the state of its post-translational modifications may be unknown~\cite{mcnaughton_2021}. 

\emph{Latent variable models (LVMs)} are particularly useful for representing biological pathways with partially known topology or missing measurements of pathway components~\citep{durbin_1998,kondofersky_2015,ernst_2007, stjohn_2019,shojaie_2009}.
LVMs are probabilistic models of a joint distribution on a set of observed and unobserved variables. A broad class of LVMs have a directed acyclic graphical (DAG) structure.
LVM-based estimation of a causal query proceeds by removing edges in the DAG that point to the target of intervention. Trained on observational data once, an LVM can estimate multiple causal queries corresponding to multiple mutilated versions of the original DAG.

There currently exists some controversy as to whether LVM-based  estimation of causal queries is accurate. One argument against this approach is that the parameters of the LVM may not be uniquely identified from the observed data~\citep{Shpitser_Evans_Richardson_Robins_2014}. Another argument is that the number of latent variables may be misspecified~\citep{shpitser_2012}. 
As a result, currently accepted approaches to LVM-based causal query estimation are limited to LVMs with specialized structural properties, such as the existence of proxy variables~\citep{louizos2017causal, kuroki2014measurement}, or the presence of multiple causes~\citep{ wang2019blessings}. 
The latter approach, although scalable to a large number of variables, is not correct in general and requires strong parametric assumptions~\citep{d2019multi}. Since biomolecular pathways have complex and diverse topology, are frequently large-scale, and have many (possibly unknown) latent variables, the controversy has so far limited the use of LVM for causal inference in this context.

In this manuscript, we argue that LVM-based estimators of causal queries are in fact accurate when the queries are identifiable according to Pearl's do-calculus, and describe a simple and practical algorithm for its estimation. We  show that the estimated probability distribution associated with the causal query converges to the true distribution, and that the estimate of its expected value is consistent.
This holds even when the parameters of the model are not uniquely identified, or when the true number of the latent variables is unknown.

We showcase the breadth of applicability, and the practical utility of LVM-based estimation of identifiable causal queries in four synthetic and two experimental case studies of biomolecular pathways. The case studies demonstrate the accuracy of the estimated causal effects, even when some pathway components are not experimentally quantified or unknown, and when the parametric assumptions only approximately represent the true data generating process. The case studies also demonstrate that the proposed approach  expands the use of causal inference to pathways where the existing alternative approaches do not apply, and enables the estimation of multiple causal queries from a single trained model.

%%%%%%%%%%%%%%%%%%%%%%%%%%%%%%%%%%%%%%%%%%%%%%%%%
\section{Background}
\subsection{Notation}
Let $\mathbf{V} = \{V_1,...,V_J\}$ be a set of observed random variables, and $\mathbf{U}=\{U_1,...,U_L\}$ be a set of latent variables. 
Let $v_i$ be an instance of  $V_i$,  and $\mathbf{v}= \{v_1,\ldots, v_j\}$ an instance of  $\mathbf{V}$.
Let $P(v_1, ..., v_j)$ be the joint probability distribution of the event $\mathbf{V} = \mathbf{v}$, and let
$P(V_i = v_i | V_j = v_j)$ be the conditional probability distribution for the event $V_i = v_i$ given $V_j = v_j$.
Denote $P(\mathbf{U})$ the prior distribution over all the latent variables, and $P(\mathbf{U}|\left\{\mathbf{v}_i\right\}_{i=1}^N)$ the posterior distribution over all latent variables $\mathbf{U}$ given $N$ observations of $\mathbf{V}$.
In this manuscript, we simplify the notation for the marginalized joint distribution $\int_\mathbf{u}P(\mathbf{U},\mathbf{V})d\mathbf{u}$ as $P(\mathbf{V})$. 
Let $G$ be a DAG with nodes $\mathbf{V} \cup \mathbf{U}$, where $\text{Pa}(V_j)$ denotes the parents of a node $V_j$ in $G$.
The joint distribution between variables $\mathbf{V} \cup \mathbf{U}$ in DAG $G$ is formulated as, $P(\mathbf{U},\mathbf{V})=\prod_{j=1}^JP(V_j|Pa(V_j))\prod_{l=1}^LP(U_l|Pa(U_l))$.
%%%%%%%%%
\subsection{Latent variable models}
A \textbf{latent variable model} (LVM) is a probability distribution over two sets of variables $\mathbf{V}$, $\mathbf{U}$, where $\mathbf{V}$ are observed at the learning time, and $\mathbf{U}$ are not observed. 
LVMs are generative, in the sense that they allow us to sample from the  joint distribution of all the variables.

A \textbf{causal LVM} $\mathcal{M}$ is an LVM with DAG structure where  $\text{Pa}(V_i)$ are interpreted as {\it direct causes} of $V_i$. 
In Bayesian framework, parameter vector $\mathbf{\theta}$ of the causal LVM are assigned prior probability distributions, and are absorbed into the set of latent variables denoted by $\mathbf{\theta}\subseteq\mathbf{U}$. 

\begin{figure}[t]
\begin{center}
\begin{tabular}{ccc}
\includegraphics[scale=0.2]{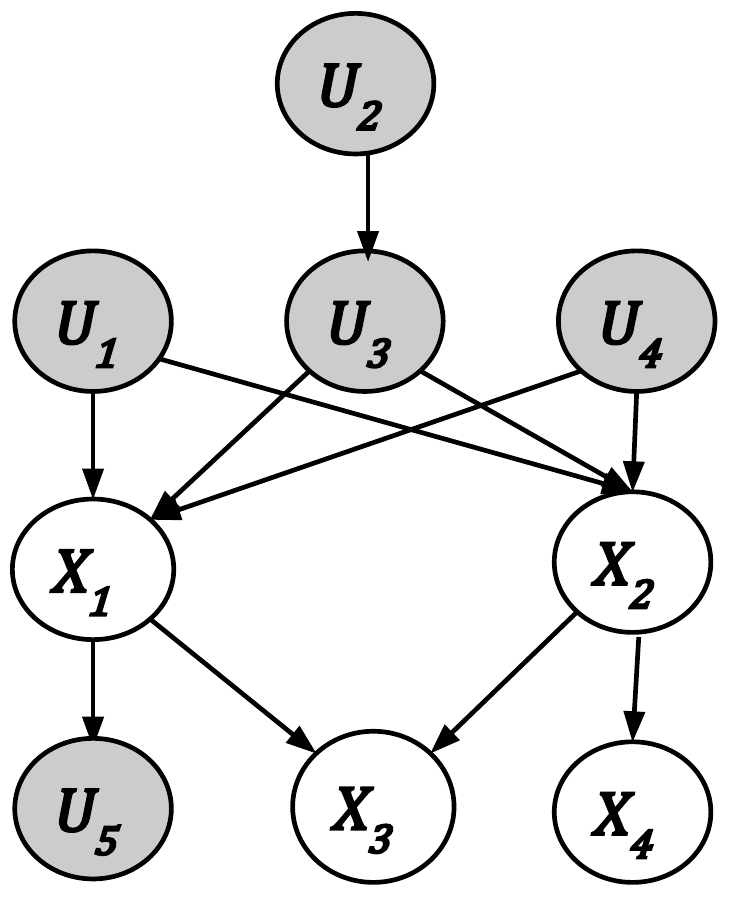} &
\includegraphics[scale=0.2]{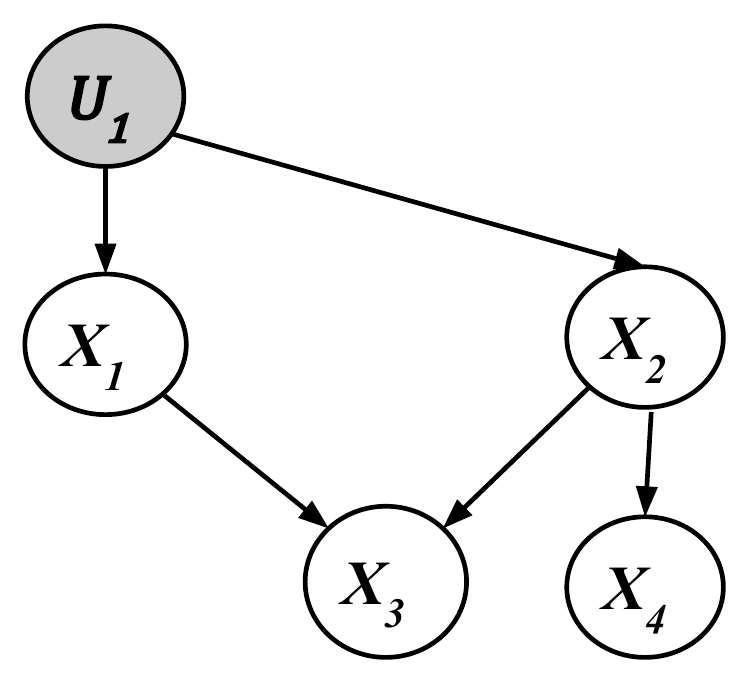} &
\includegraphics[scale=0.2]{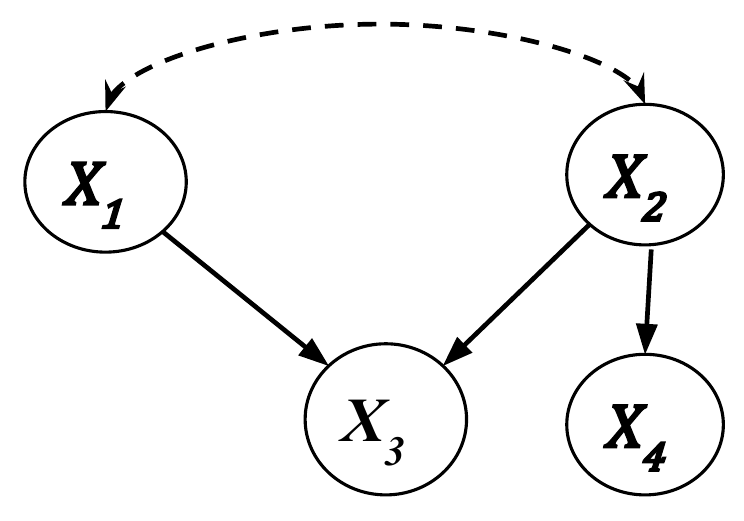}\\
     (a)   & (b) & (c)
\end{tabular}
\end{center}
    \caption{\small  
(a) An LVM with 4 observed (white) and 5 latent (dark grey) variables.
(b) A different LVM with 1 latent variable.
(c) An ADMG representing both (a) and (b). 
\label{fig:DAGToADMG}}
\end{figure}

Given a causal LVM with a DAG $G$, observed variables $\mathbf{V}$, and latent variables $\mathbf{U}$, \citep{evans2016graphs} compactly represents LVMs with many latent variables by an LVM with a single latent variable between each pair of observed variables, according to the following rules:
\begin{enumerate}
%\vspace{-5mm}
    \item Remove latent variables with no children. 
%\vspace{-2mm}
    \item Remove a latent variable $U$ with observable parents by connecting all the parents of $U$ to its children. 
%\vspace{-2mm}
    \item If $U, W$ are latent variables with $children(W ) \subseteq children(U)$, then remove $W$.
\end{enumerate}
%\vspace{-5mm}
\figref{fig:DAGToADMG} (a) illustrates a causal LVM with many latent variables, and 
\figref{fig:DAGToADMG} (b) a causal LVM obtained from (a) by applying the simplification.
\figref{fig:DAGToADMG} (c) is an \textbf{acyclic directed mixed graph (ADMG)}~\citep{richardson_2017} representing both \figref{fig:DAGToADMG} (a) and (b). It shows the existence of latent variables between $X_1$ and $X_2$ by a dashed bi-directed edge.

{\bf Inference algorithms} \cite{10.5555/1162264} sample from the posterior distribution $P(\mathbf{U} | \left\{\mathbf{v}_i\right\}_{i=1}^{N})$ of latent variables in the ADMG, including the parameters $\theta$, given $N$ observations of $\mathbf{V}$. In particular, exact algorithms such as Hamiltonian Monte Carlo (HMC) \citep{ girolami2011riemann} guarantee asymptotically exact samples, but are computationally expensive \citep{robert2013monte}.
Approximate probabilistic inference algorithms such as variational inference \citep{blei2017variational} trade off accuracy for speed by searching with gradient descent a parameterized family of functions that approximate the target distribution. A trained causal LVM $\hat{\mathcal{M}}$ is an LVM where posterior distributions of the parameters are learned with an inference algorithm. Many packages such as PyStan~\citep{van2013python}, or pyro~\citep{bingham2019pyro}  in Python, or RStan~\citep{rstanInterface} in R take as input an LVM and output a trained LVM.
\subsection{Causal query identification}
\label{CI}

Frequently, we are interested in an \textbf{intervention} on a set of target variables $\mathbf{X}\subseteq \mathbf{V}$ which fixes a set of variables $\mathbf{X}$ to constant  values $\mathbf{x'}$ (denoted $do(\mathbf{X} = \mathbf{x'})$, shortened to $do(\mathbf{x'})$), and makes it independent of its causes~\citep{spirtes2000causation, eberhardt2007interventions}. {\bf Graph mutilation} in a causal LVM simulates an intervention.
It severs the edges incoming to the target nodes, and fixes each node $X\in\mathbf{X}$ to its intervention value $x' \in \mathbf{x'}$~\citep{koller2009probabilistic}, producing a graph that we denote  $G_{\bar{\mathbf{X}}}$. Denote $P_{G_{\bar{\mathbf{X}}}}(\mathbf{v})$ the probability distribution encoded by ${G_{\bar{\mathbf{X}}}}$. Denote $\mathcal{M}_{\bar{\mathbf{X}}}$ the causal LVM with  structure $G_{\bar{\mathbf{X}}}$ (the subscript ${\bar{\mathbf{X}}}$ in this notation distinguishes the intervened model from the original model). Denote $P_{\mathcal{M}_{\bar{\mathbf{X}}}}(\mathbf{v})$ the probability distribution, and $E_{\mathcal{M}_{\bar{\mathbf{X}}}}[\mathbf{v}]$ the expected value of the variables $\mathbf{v}$ in the intervened model $\mathcal{M}_{\bar{\mathbf{X}}}$.

A \textbf{causal query} $Q_{\mathbf{X}}$ with respect to a causal LVM $\mathcal{M}$ is a probabilistic query that conditions a set of outcomes $\mathbf{Y}\subseteq\mathbf{V}\backslash{\mathbf{X}}$ on a set of interventions, such as $Q_{\mathbf{X}}=P_{{\mathcal{M}}_{\bar{X}}}(\mathbf{Y}| do(\mathbf{x'}))$ or $Q_{\mathbf{X}}=E_{{\mathcal{M}}_{\bar{X}}}[\mathbf{Y} | do(\mathbf{x'})]$.  
To denote the distribution of the outcome variable obtained from a mutilated model that was trained on pre-interventional data, we use counterfactual subscript notation $\mathbf{Y}_{do(\mathbf{x}')}\sim P(\mathbf{Y}_{do(\mathbf{x}')}|\{x_i, y_i\}_{i=1}^N)$.

A causal query $Q_{\mathbf{X}}$ is {\bf identifiable} with respect to $P(\mathbf{V})$ and an ADMG $A$, if all LVMs that project onto $A$ and agree on $P(\mathbf{v})$ also agree on the value of $Q_{\mathbf{X}}$ \citep{shpitser2008complete}. A causal query is identifiable if it satisfies the back-door or the front-door criteria~\citep{pearl2009causality}. The back-door and the front-door criteria rely on the following concepts of graphical modeling. In a DAG $G$ there is a \textbf{path} between $V_i$ and $V_j$, if there is a sequence of edges connecting $V_i$ to $V_j$. A variable is a \textbf{collider} when both edges adjacent to the variable on the path point into it. A path is \textbf{blocked} if we observe the value of a non-collider on that path or we do not observe the value of a collider.

The \textbf{back-door criterion} \citep{pearl2009causality} holds for $X, Y\in\mathbf{V}$ in ADMG $A$ if there is no path from $X$ to $Y$ consisting of bi-directed edges, and there exists a set $\mathbf{Z}\subseteq \mathbf{V}\backslash\{X,Y\}$ such that no node is a descendant of $X$, and $\mathbf{Z}$ blocks every path between $X$ and $Y$ that contains an arrow into $X$~\citep{pearl2009causality}.
If a set of variables $\mathbf{Z}$ satisfies the back-door criterion relative to $(X,Y)$, then the causal effect of $X$ on $Y$ is identifiable and is given by $P(Y|do(x')) = \sum_{z} P(Y | x',z) P(z)$.
The \textbf{front-door criterion}  \citep{pearl2009causality} holds when there is an unobserved confounder, but there exists a mediator between cause and effect that is shielded from confounding~\citep{pearl2009causality,pearl1993bayesian,pearl1995causal}. If a set of variables $\mathbf{Z}$ satisfies the front-door criterion relative to
$(X, Y )$ and if $P(x, z) > 0$, then the causal effect of $X$ on $Y$ is
identifiable and is given by the formula, $P(Y|do(x')) = \sum_{z} P(z | x') \sum_{x} P(Y | x, z) P(x)$.
For example, neither the back-door nor the front-door criterion hold in \figref{fig:DAGMutDAG} (a) but the front-door criterion holds in \figref{fig:DAGMutDAG}(b). The back-door and front-door criteria are sufficient but not necessary for causal identifiability. 
%%%%%%%%%%%%%%%%%%%%%%%%%%%%%%%%%%%%%

The {\bf do-calculus} is comprised of three rules for symbolically manipulating interventional and observational joint distributions. 
Let $\mathbf{X}$, $\mathbf{Y}$, $\mathbf{Z}$, and $\mathbf{W}$ be disjoint sets of variables in the joint distribution entailed by ADMG G. 
Let $G_{\overline{\mathbf{X}}}$ denote the graph produced by mutilating $G$ such that all incoming edges to $\mathbf{X}$ are removed. 
Similarly $G_{\underline{\mathbf{Z}}}$ is the graph created
when $G$ is mutilated by removing all outgoing edges from $\mathbf{Z}$. The 3 rules of do-calculus are as follows.

\noindent \textbf{1:} $P(\mathbf{Y}|do(\mathbf{x}), \mathbf{z}, \mathbf{w}) = P(\mathbf{Y}|do( \mathbf{x}), \mathbf{w})$ if 
$(\mathbf{Y} \indep \mathbf{Z} | \mathbf{X}, \mathbf{W}) G_{\overline{\mathbf{X}}}$

\noindent \textbf{2:} $P(\mathbf{Y}|do(\mathbf{x},  \mathbf{z}), \mathbf{w}) = P(\mathbf{Y}|do( \mathbf{x}), \mathbf{z}, \mathbf{w})$ if $(\mathbf{Y} \indep \mathbf{Z}|\mathbf{X}, \mathbf{W}) G_{\mathbf{X} \underline{\mathbf{Z}}}$

\noindent \textbf{3:} $P(\mathbf{Y}|do(\mathbf{x}, \mathbf{z}), \mathbf{w}) = P(\mathbf{Y}|do( \mathbf{x}), \mathbf{w})$ if $(\mathbf{Y} \indep \mathbf{Z}|\mathbf{X}, \mathbf{W})G_{\overline{\mathbf{X}}, \overline{ \mathbf{Z}(\mathbf{W})}}$
Here, $\mathbf{Z}(\mathbf{W})$ is the subset of nodes in $\mathbf{Z}$ that are not ancestors of any node in $\mathbf{W}$. The do-calculus
rules are complete \cite{10.5555/3020419.3020446, shpitser2006identification}, meaning if a causal query is
identifiable, then it can be derived using these three rules.

A causal query containing a $do()$ operator is identifiable in a given ADMG if the do-calculus transforms it into an equivalent $do$-free estimand.
The do-calculus estimands are non-parametric, in the sense that they do not impose constraints on $P(\mathbf{x})$. 
Any causal query in an ADMG identifiable by the do-calculus is also identifiable in every causal LVM that projects onto that ADMG~\citep{richardson_2017}.

%\subsection{Implementations}
Several sound and complete algorithms take as input an ADMG and a causal query, and determine whether the query is identifiable according to the do-calculus~\citep{ richardson_2017,shpitser2008complete}. These algorithms have polynomial time complexity \cite{galles2013testing}.

\subsection{Causal query estimation}

For queries of a form of $P_{\mathcal{M}_{\bar{X}}}(\mathbf{Y}|do(\mathbf{x'}))$, a desirable property of the estimator is the convergence of the estimated probability distribution to the true probability distribution. 
For queries of a form of $E_{\mathcal{M}_{\bar{X}}}[\mathbf{Y}|do(\mathbf{x'})]$, a desirable property of the estimator is consistency. 
An estimator of $E_{\mathcal{M}_{\bar{X}}}[\mathbf{Y} | do(\mathbf{x'})]$ is \textbf{consistent} if, as the number of data points used to estimate the query tends to infinity, the sequence of the estimates converges in probability to its expected value.

Several non-LVM approaches for estimating causal queries with these desirable properties exist such as semi-parametric primal IPW (PIPW), dual IPW (DIPW), Nested IPW and augmented nested IPW~\citep{bhattacharya2020semiparametric}.
They are all implemented and well-documented in Ananke~\cite{bhattacharya2020semiparametric}.
Unfortunately, these approaches derive a separate statistical estimand for each causal query anew~\citep{pearl_2019}. In addition, they are limited to causal queries with one cause and one effect, and the cause must be binary-valued.
This has limited, the scope of their applicability in systems biology where one is often interested in the simultaneous effect of multiple cause on one or multiple effects and the variables are not always discrete.
Other approaches such as (WERM-ID)~\cite{jung2020learning} and double/debiased machine learning (DML)~\cite{jung2021estimating} proposed estimators for any identifiable query but are inadequate in large data regimes where it is computationally expensive to train a new estimator for each query of interest. The implementations for these approaches are unavailable for the public.

In this manuscript, we advocate for the explicit use of LVMs for causal query estimation in presence of latent variables when causal queries contain multiple-causes, non-discrete cause(s), or multiple effects, as these are common in biology. We demonstrate that if the graph topology of an LVM correctly reflects the true underlying causal structure of the observed variables, and if the causal query of interest is identifiable according to Pearl's do-calculus, then LVM-based estimators have the desired properties.

%%%%%%%%%%%%%%%%%%%%%%%%%%%%%%%%%%%%%%%%%%%%%%%%%%
\begin{small}
\begin{algorithm}[ht!]
\small
\caption{Estimation of an identifiable causal query}
\label{alg:cap}
\DontPrintSemicolon
\SetAlgoLined
\KwInput{
$~~~~~~~$ $\hat{\mathcal{M}}$, a causal LVM trained on observational data with an exact inference algorithm\\
$~~~~~~~~~~~~~~~~~~~~~$ $\mathbf{x}'\subseteq \mathbf{v}$, target values of the intervention\\
$~~~~~~~~~~~~~~~~~~~~~$ $\mathbf{Y}\subseteq \mathbf{V}$, effects of the intervention\\
$~~~~~~~~~~~~~~~~~~~~~$ $Q_{\mathbf{X}}=
P_{\mathcal{M}_{\bar{X}}}(\mathbf{Y}| do(\mathbf{x}^{\prime}))$ or
$E_{\mathcal{M}_{\bar{X}}}[\mathbf{Y} | do(\mathbf{x}^{\prime})]$ query\\
}
\KwParameter{ 
$S$, \# of samples from the posterior distribution \\
$~~~~~~~~~~~~~~~~~~~~$ $L$, \# of samples for each variable\\
}
\KwOutput{$~~~~~$ $\hat{P}_{\hat{\mathcal{M}}_{\bar{X}}}(\mathbf{Y}|do(\mathbf{x}'))$ or $\hat{E}_{\hat{\mathcal{M}}_{\bar{X}}}[\mathbf{Y}|do(\mathbf{x}')]$}

\vspace{0.2cm}
\hrule 
\vspace{0.2cm}
{\color{brown} \ttfamily \scriptsize //Check identifiability of $Q_{\mathbf{X}}$}

\textbf{if} $Q_{\mathbf{X}}$  \textit{is not identifiable}  \textbf{then raise }  \text{not identifiable error} \label{checkIdentNot}

Set $\mathbf{X} = \mathbf{x}'$        \label{e}

Create $\hat{\mathcal{M}}_{\bar{\mathbf{X}}}$, the mutilated model   \label{mutilate}

\For{$s$ in 1:$S$}  
 {
 	Sample $\theta_s \sim P_{\hat{\mathcal{M}}_{\bar{\mathbf{X}}}} (\theta | \{\mathbf{v}_i\}_{i=1}^N)$  \label{sampleTheta}
 }

\For{ $W$ in topological-sort$(\{\mathbf{U} \cup \mathbf{V}\} )$} 
{
	Sample $w_{s} \sim P_{\hat{\mathcal{M}}_{\bar{\mathbf{X}}}}(W | Pa(W); \theta_s)$ $L$ times \label{sampleVar}
	
	Collect $ \mathbf{y}_{s} \subseteq\mathbf{w}_s$  \label{f}
}

\textbf{Return} density$\left(\{\mathbf{y}_s\}_{s=1}^S\right)$ or $\frac{1}{S}\sum_{s=1}^S \mathbf{y}_s$ \label{calcQuery}

\end{algorithm}
\end{small} 
%%%%%%%%%%%%%%%%%%%%%%%%%%%%%%%%%%%%%%%%%%%%%%%%%%%%%
%%%%%%%%%%%%%%%%%%%%%%%%%%%%%%%%%%%%
%%%%%%%%%%%%%%%%%%%%%%%%%%%%%%%%%%%%%%%%%%%%%
%%%%%%%%%%%%%%%%%%%%%%%%%%%%%%%%%%%%%%%%%%%%%%%%%%%%%%%%%%%%%%%%%%
%%%%%%%%%%%%%%%%%%%%%%%%%%%%%%%%%%%%%%%%%%%%%%%%%%%%%
%%%%%%%%%%%%%%%%%%%%%%%%%%%%%%%%%%%%
%\begin{methods}
\section{Methods}
\label{methods}

\subsection{Contribution of this work}
In this manuscript, we propose a simple and practical algorithm (Algorithm~\ref{alg:cap}) for LVM-based causal query estimation. The algorithm takes as inputs a causal query of interest in the form of the distribution over the effect $Y$ given an intervention on the cause $X$, i.e. $\mathbf{Y}_{do(\mathbf{x'})}\sim P_{\mathcal{M}_{\bar{X}}}(\mathbf{Y}|do(\mathbf{x}'))$, 
or in the form of the expected value of this distribution, i.e. $E_{\mathcal{M}_{\bar{X}}}[\mathbf{Y}|do(\mathbf{x}')]$, target values of the intervention, effects of the intervention, and an LVM with known DAG or ADMG structure that is trained on observational data. The output of the algorithm is the estimate of the causal query of interest, i.e. $\hat{P}_{\hat{\mathcal{M}}_{\bar{X}}}(\mathbf{Y}|do(\mathbf{x}'))$ or $\hat{E}_{\hat{\mathcal{M}}_{\bar{X}}}[\mathbf{Y}|do(\mathbf{x}')]$. 

The algorithm first determines whether the causal query of interest is identifiable according to Pearl's do-calculus (\algref{checkIdent}). If the query is identifiable, Algorithm~\ref{alg:cap}  proceeds with its estimation. 
We take a Bayesian viewpoint~\citep{lattimore_2019, lattimore_2019a}, and follow the abduction, action, prediction paradigm~\citep{pearl2009causality}. Abduction estimates the posterior distribution over the latent variables (including the model parameters) given the training data. A trained LVM, including these posterior distributions, is an input to Algorithm~\ref{alg:cap}. 
Action fixes the values of the intervened variables (\algref{e}) and breaks the relationship of the intervened variables to their parents (\algref{mutilate}). 
Prediction samples the parameters from their posterior distributions (\algref{sampleTheta}), and then samples from each variable given its parents (\algref{sampleVar}) until we are ready to estimate the causal query (\algref{calcQuery}). 
Thus the estimator can be thought of as a posterior predictive statistic over the marginal of the parameters.

The algorithm takes as input a trained LVM. In particular, it can take a trained LVM with continuous distributions, and multiple causes and effects, where non-parametric or current parametric approaches are limited. While training an LVM is NP-complete (and in practice depends on the specific LVM and on the choice of inference algorithm), it amortizes most of the computational work into this single training step. 
Given a single trained model, it can estimate an arbitrary number of queries.

%%%%%%%%%%%%%%%%%%%%%%%%%%%%%%%%%%%%%%%%%%%%%%%%%%%%%%%%%
\begin{figure}[h]
\begin{center}
\begin{tabular}{c|c}
\includegraphics[scale=0.23]{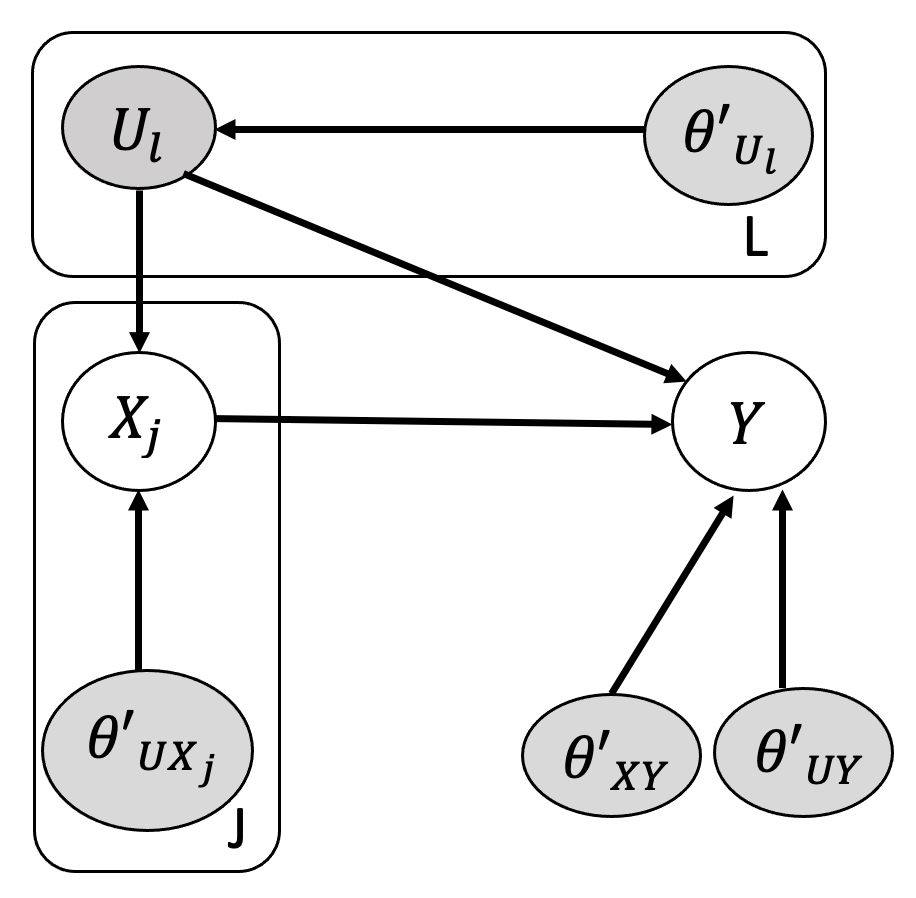} &
\includegraphics[scale=0.15]{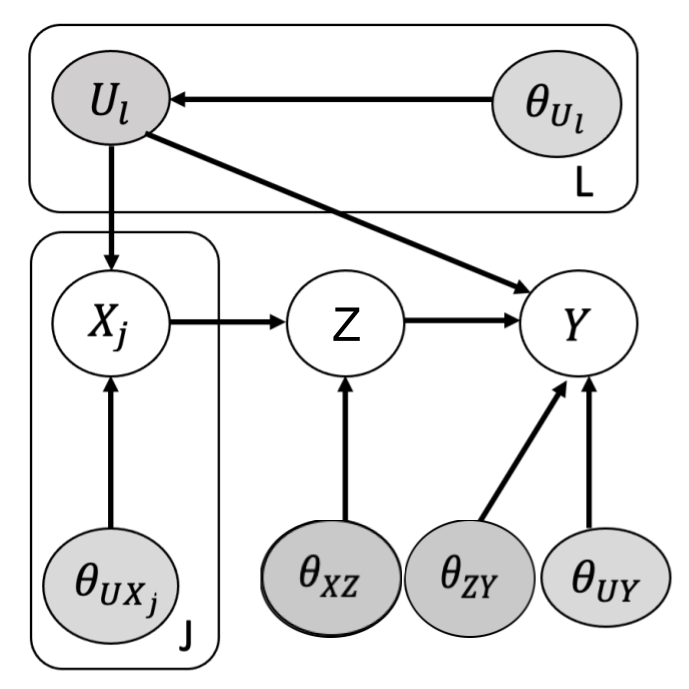}\\
(a) & (b) \\
\includegraphics[scale=0.18]{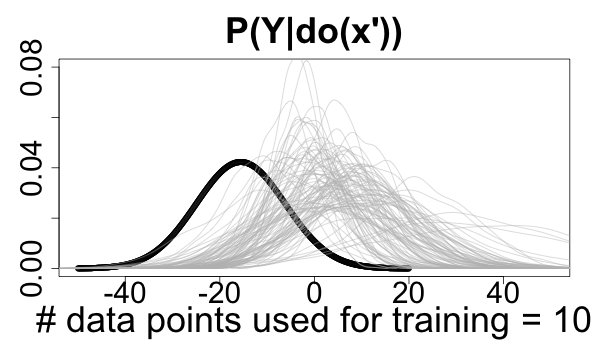} &
\includegraphics[scale=0.18]{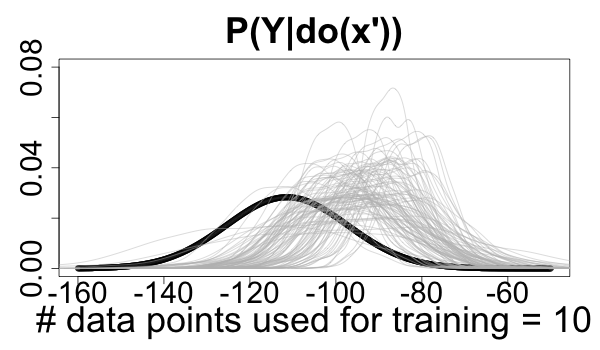} \\
(c) & (d) \\
%\includegraphics[scale=0.18]{img/example1Density2.png} &
%\includegraphics[scale=0.18]{img/example2Density2.png} \\
%(e) & (f) \\
\includegraphics[scale=0.18]{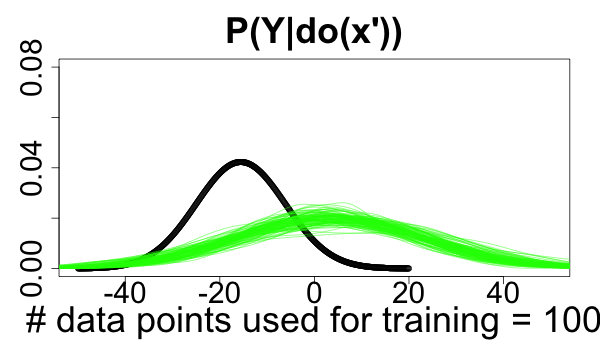} &
\includegraphics[scale=0.18]{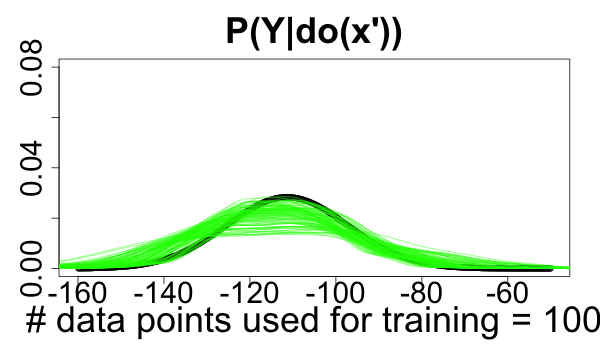} \\
(e) & (f)
\end{tabular}
\end{center}
    \caption{\small  
\textbf{The estimates of a non-identifiable causal query $P_{\mathcal{M}_{\bar{X}}}(Y|do(\mathbf{x}'))$ fail to converge to the true distribution as number of data points used to train the LVM increases (left column). The estimates of an identifiable causal query converges to the true distribution (right column).} 
(a) An LVM where $P_{\mathcal{M}_{\bar{X}}}(Y|do(\mathbf{x}'))$ is not non-parametrically identified. Boxes indicate sets of variables with the same structure. Circular white/gray nodes are observed/latent variables. $\theta'$ are model parameters. Each parameter such as $\theta'_U$ has a prior distribution, e.g. $\theta'_U \sim P(q_{\theta'_U})$, where $q_{\theta'_U}$ is a hyperparameter. 
(b) As in (a), but in this case $P_{\mathcal{M}_{\bar{X}}}(Y|do(\mathbf{x}'))$ is non-parametrically identified. 
(c,e) relate to (a). Black curve estimates the true distribution $P_{\mathcal{M}_{\bar{X}}}(Y|do(\mathbf{x}');\theta)$,\ with $\theta$ used to generate interventional data. After training the LVM on $N=10, 100$ observational data points, each gray/green curve estimate $P_{\hat{\mathcal{M}}_{\bar{X}}}(Y_{do(\mathbf{x}')};\{x_i, y_i\}_{i=1}^N, \theta)$ for each sampled $\theta$. The curves do not approach the true distribution as number of data points increases. 
(d,f) relate to (b). The curves converge to the true distribution as the number of data points increases.  
\label{fig:DAGMutDAG}
}
\end{figure}
%%%%%%%%%%%%%%%%%%%%%%%%%%%%%%%%%%%%%%%%%%%%%%%%%%%%
\subsection{Convergence and consistency of the estimator in Algorithm~\ref{alg:cap} in correctly specified LVMs}

\textbf{Motivating examples} We illustrate the practical application of Algorithm~\ref{alg:cap} in the special case of the  LVM in \figref{fig:DAGMutDAG}(a). It represents a situation where, e.g. a protein product of gene $X$ affects  gene $Y$,  while  both  are  under  regulation of the same transcription factor(s) and/or enhancer(s).
The causal query $P_{\mathcal{M}_{\bar{X}}}(Y|do(x'))$ is not identifiable, and we show empirically that its LVM-based estimator is biased (\figref{fig:DAGMutDAG}(c,e)).

Extending the causal LVM with a mediator $Z$ in~\figref{fig:DAGMutDAG}(b) makes the query identifiable according to the front-door criterion. 
This pattern occurs frequently in transcriptional cascades which involve multiple steps, or signaling pathways in which $Y$ is not a direct substrate of $X$.
We show empirically that the estimate of $P_{\mathcal{M}_{\bar{X}}}(Y|do(x'))$ converges to the true distribution (\figref{fig:DAGMutDAG}(d,f)).

\textbf{Empirical example 1: \figref{fig:DAGMutDAG}(a)} 
Assume a model $\mathcal{M}$:
$U := \theta'_U;
X := U \theta'_{UX} + \theta'_{X};
Y := X \theta'_{XY} + U \theta'_{UY} + \theta'_{Y}$
where  $\theta'_{X}\sim N(\mu'_X, \sigma'_X), \theta'_Y\sim N(\mu'_Y, \sigma'_Y), \theta'_U\sim N(\mu'_U, \sigma'_U)$ and a non-identifiable causal query of $P_{\mathcal{M}_{\bar{X}}}(Y|do(x'))$.
We generated observational data with $N=10,100$ samples from the likelihood with a randomly chosen vector of true values of $\theta$. 
The true $P_{\mathcal{M}_{\bar{X}}}(Y|do(x'); \theta)$ was estimated with Algorithm~\ref{alg:cap}, where \algref{sampleTheta} was substituted by the true values of $\theta$ (black curves in \figref{fig:DAGMutDAG}(c,e)).

To learn a model $\hat{\mathcal{M}}$ from this training data, we assumed a Gaussian prior on the parameters: 
$\mu_U',\mu_X', \mu_Y', \sigma_U',\sigma_X',\sigma_Y', \theta'_{UX} \sim N(0,1)$ and  $\theta'_{XY}, \theta'_{UY}  \sim N(0,10)$, and trained the model with HMC. 
Thin lines in \figref{fig:DAGMutDAG}(c,e) estimate $P_{\hat{\mathcal{M}}_{x'}}(Y|do(x'), \{x_i, y_i\}_{i=1}^N, \theta)$ for each sampled $\theta$ (\algref{sampleVar}). As $N$ increases, the distributions became less diverse,  but did not approach the ground truth.

\textbf{Empirical example 2: \figref{fig:DAGMutDAG}(b)} 
Expanding the previous example with a mediator $Z$, we assume a model
$U := \theta_{U}$,
$X := U \theta_{UX} + \theta_{X}$,
$Z := X \theta_{XZ} + \theta_{Z}$,
$Y := Z \theta_{ZY} + U \theta_{UY} + \theta_{Y}$
where,
$\theta_U\sim N(\mu_U,\sigma_U)$,
$\theta_X\sim N(\mu_X,\sigma_X),
\theta_Y\sim N(\mu_Y,\sigma_Y),
\theta_Z\sim N(\mu_Z,\sigma_Z)$.

With this expansion, the causal query $P_{\mathcal{M}_{\bar{X}}}(Y|do(x'))$ becomes identifiable. Repeating the same analysis, \figref{fig:DAGMutDAG}(d,f)
show that, as $N$ increased, the distributions converged to the ground truth. 
The analytical proof of this empirical result for multivariate $U$ and $X$ can be found in Supplementary Materials.

The following Lemma 1 proves the empirical results for any arbitrary distribution.

\medskip \textbf{Lemma 1} 
{\it Consider the LVM in \figref{fig:DAGMutDAG} (b) with a DAG $G$. $\mathbf X$, $Z$, and $Y$ are observed and $\mathbf U$ are latent. The front-door adjustment estimand of the query $P(Y|do(\mathbf{x}'))$ is equivalent to the estimand of that query in the mutilated LVM.}

\begin{proof}
Consider a mutilated version of $G$, $G_{\bar{\mathbf{X}}}$, where all the incoming edges to $\mathbf{X}$ are removed. A causal query $P(Y|do(\mathbf{x}'))$ transforms $P(.)$ into a distribution 
%$P_{{G_{\bar{\mathbf X}}}}(.)$,
$P_{\bar{\mathbf{X}}}(.)$, and $P(Y|do(\mathbf x')) = P_{\bar{\mathbf{X}}}(Y|\mathbf x')$. Hence,
\begin{align}
&P(Y | do(\mathbf{x}')) = 
P_{{\bar{X}}}(Y|\mathbf{x}')
= \int_{\mathbf{u}, z} P_{{\bar{\mathbf X}}}(Y, \mathbf{u}, z|\mathbf{x}')d\mathbf{u} dz
\nonumber \\
=&\int_{z} \left(\int_{\mathbf{u}} P_{{{\bar{\mathbf X}}}}(Y|\mathbf{u}, z, \mathbf{x}') P_{{{\bar{\mathbf X}}}}(\mathbf{u} | z, \mathbf{x}')   d\mathbf{u}\right) P_{{{\bar{\mathbf X}}}}(z | \mathbf{x}')  dz  \nonumber\\
=& \int_{z} P_{{{\bar{\mathbf X}}}}(Y|z) P_{{{\bar{\mathbf X}}}}(z | \mathbf{x}')  dz \nonumber\\
=& \int_{z} P(Y|do(z)) P(z | \mathbf{x}')  dz \label{ffee}\\
=& \int_{z}  \left (\int_{\mathbf{x}} P(Y|z, \mathbf{x}) P(\mathbf{x} )  d\mathbf{x} \right ) P(z | \mathbf{x}') \label{mgraph} dz
\end{align}

\eqref{ffee} holds because in $G_{\bar{\mathbf{X}}}$, $Y$ is independent from $\mathbf{X}$ given $Z$.
Since $P_{\bar{\mathbf{X}}}(z | \mathbf{x}')$ is unaffected by the mutilation of $G$, $P_{\bar{\mathbf {X}}}(z | \mathbf{x}') = P_{G}(z | \mathbf{x}')$.  
\eqref{mgraph} follows from the back-door path between $Y$ and $Z$ in $G$. The expression on the right-hand side of \eqref{mgraph} is the estimand for $P(Y|do(\mathbf x'))$ derived from the do-calculus front-door adjustment formula.  \qedsymbol 
\end{proof}

The following Theorem proves that in general, for any LVM topology, any set of parametric distributions, and any identifiable causal query, Algorithm~\ref{alg:cap} accurately estimates causal queries in an LVM that correctly reflects the true underlying causal structure.

\medskip \textbf{Theorem 1} {\it  
Consider a causal LVM $\mathcal{M}$, which includes the true likelihood that generated the observational data. Consider a causal query 
$Q_{\mathbf{X}}=P_{\mathcal{M}_{\bar{X}}}(\mathbf{Y}| do(\mathbf{x'}))$ or $Q_{\mathbf{X}}=E_{\mathcal{M}_{\bar{X}}}[\mathbf{Y} | do(\mathbf{x'})]$, identifiable according to the do-calculus with respect to $\mathcal{M}$.
When estimating the causal query as in Algorithm~\ref{alg:cap}, the estimate $\hat{P}_{\hat{\mathcal{M}}_{\bar{X}}}(\mathbf{Y}| do(\mathbf{x'}))$ converges to the true distribution, and the estimator $\hat{E}_{\hat{\mathcal{M}}_{\bar{X}}}[\mathbf{Y} | do(\mathbf{x'})]$ is consistent.
}

\begin{proof}
When the ground truth parameters $\theta$ are known, samples from the likelihood $v_s \sim P(V | Pa(V), \theta)$ for all $V \in \mathbf{V}$ converge to the true joint observational distribution $\prod_{V\in\mathbf{V}}P(V|Pa(V),\theta)$ as $N\rightarrow\infty$. $N$ is the number of data points.
%%%%%%%%%%%%%%%%%%%%%%%%%%%%%%%%%%%%%%%%%%%%%%%%
\begin{figure}[H]
\begin{center}
\begin{tabular}{c}
\includegraphics[scale=0.25]{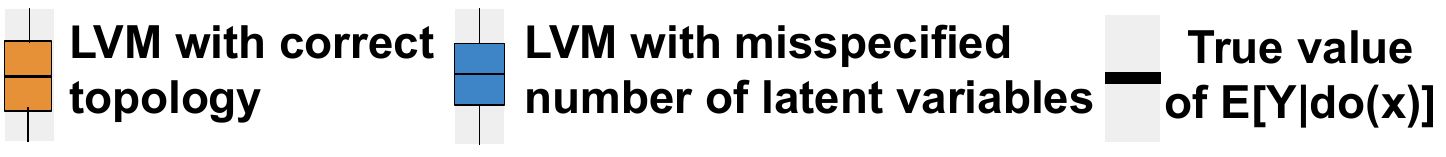}
\end{tabular}
\begin{tabular}{cc}
\includegraphics[scale=0.19]{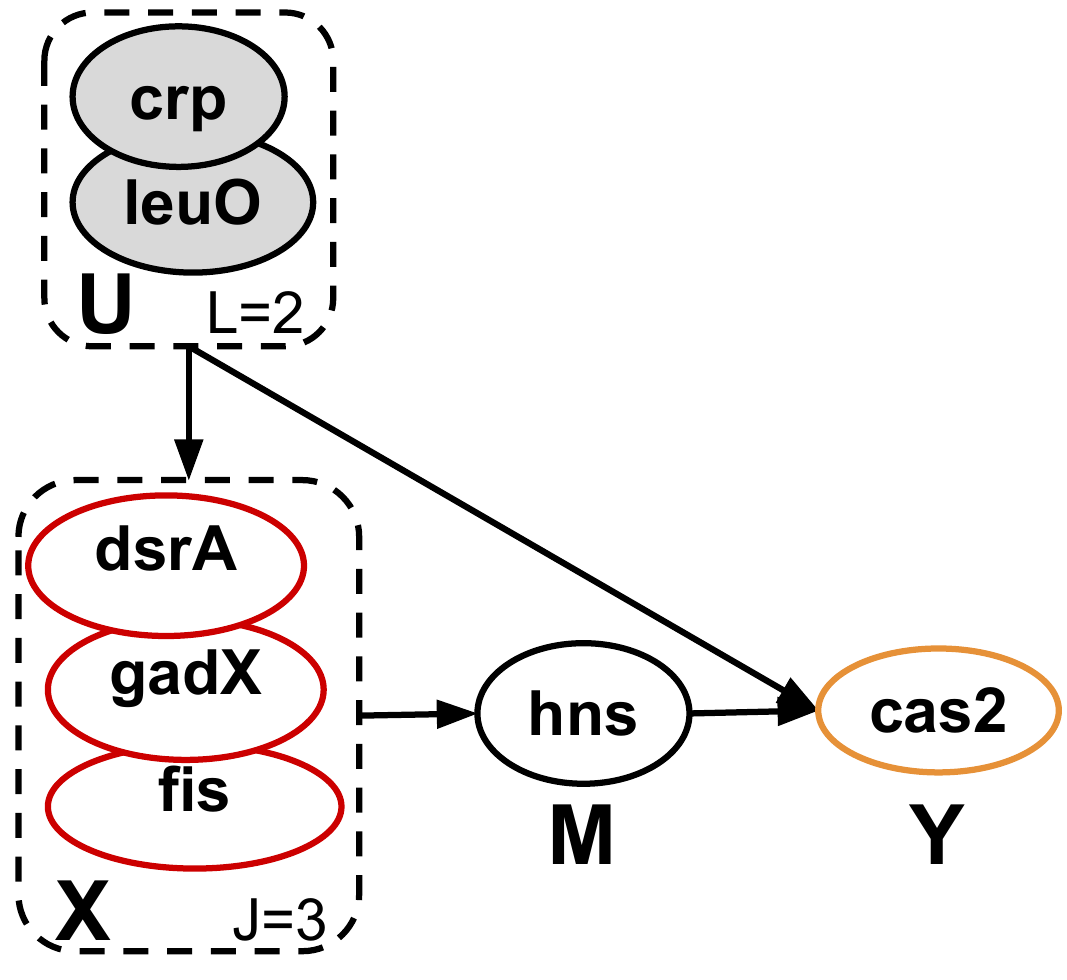}&
\includegraphics[scale=0.19]{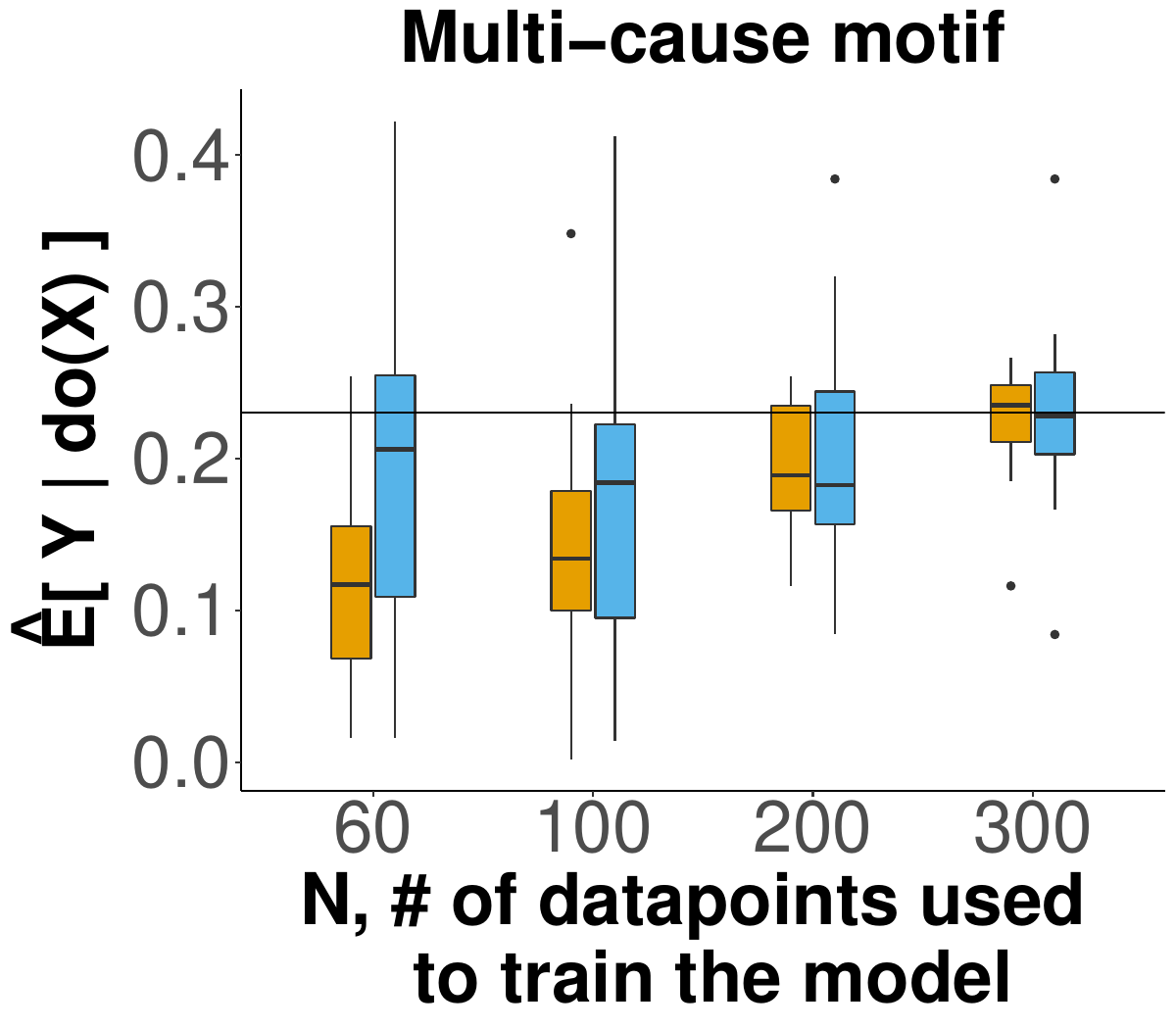} \\
(a)  Case study 1 : Multi-cause motif & (b) $\hat{E}[cas2| do(dsrA, gadX, fis)]$
\end{tabular}
\end{center}
    \caption{\small 
     \textbf{Synthetic case study 1.} Red nodes are targets of the intervention, orange nodes are the effect. gray nodes are latent.(a) The multi-cause feed-forward transcriptional regulatory network motif.
    (b) Sampling distribution of $\hat{Q}_{\mathbf{x}}=\hat{E}[cas2| do(dsrA, gadX, fis = 0)]$ over 20 observational datasets.
     \label{fig:case1}}
\end{figure}
%%%%%%%%%%%%%%%%%%%%%%%%%%%%%%%%%%%%%%%%%%%%%%%%%%
In practice parameters of the LVM are trained on observational data. If the parameters are not identifiable during training, their posterior distribution $\theta_r \sim P(\theta|\{\mathbf{v}_i\}_{i=1}^N)$ is not guaranteed to converge to the true value.
Nonetheless, samples from the observed variables $v_s\sim P(V|Pa(V), \theta_r)$, $V\in\mathbf{V}$, converge to the same true joint observational distribution $\prod_{V\in\mathbf{V}}P(V|Pa(V),\theta)$. 
For identifiable causal queries,  all parametrizations that agree on the joint observational distribution agree on the queries~\citep{shpitser2008complete}. 
Therefore, since under stability conditions exact inference algorithms provide guarantees of asymptotically exact samples, the posterior predictive distribution 
$P(\mathbf{Y}_{do(\mathbf{x}')}| \{\mathbf{v}_i\}_{i=1}^N)$ converges to the true distribution, and its expected value 
$E[\mathbf{Y}_{do(\mathbf{x}')} | \{\mathbf{v}_i\}_{i=1}^N]$ is consistent \cite{robert2013monte,BDA3}. \qedsymbol
\end{proof}

\subsection{Convergence and consistency of the estimator in Algorithm~\ref{alg:cap} in presence of miss-specified number of latent variables}

The following corollary proves that queries of the form of $E_{\mathcal{M}_{\bar{X}}}(\mathbf{Y}| do(\mathbf{x}))$ or 
$P_{\mathcal{M}_{\bar{X}}}(\mathbf{Y}| do(\mathbf{x}))$ can be accurately estimated even when  the true number of latent variables is unknown.

\medskip \textbf{Corollary 1} {\it Consider a causal LVM $\mathcal{M}$, which includes the true likelihood that generated the observational data.  Consider a class of LVMs $\mathbb{M}$ that projects on the same ADMG as $\mathcal{M}$. Consider a causal query 
$Q_{\mathbf{X}}=P_{\mathcal{M}_{\bar{X}}}(\mathbf{Y}| do(\mathbf{x'}))$ or $Q_{\mathbf{X}}=E_{\mathcal{M}_{\bar{X}}}[\mathbf{Y} | do(\mathbf{x'})]$, identifiable according to the do-calculus with respect to $\mathbb{M}$.  When estimating the causal query as in Algorithm~\ref{alg:cap}, the estimate $\hat{P}_{\hat{\mathcal{M}}_{\bar{X}}}(\mathbf{Y}| do(\mathbf{x'}))$ converges to the true distribution, and the estimate $\hat{E}_{\hat{\mathcal{M}}_{\bar{X}}}[\mathbf{Y} | do(\mathbf{x'})]$ is consistent.}

\begin{proof}
Let $\theta'$ be the parameters of $\mathcal{M}' \in \mathbb{M}$. 
Following the same logic as in proof of Theorem, the samples $v'_s 
\sim P(V | Pa(V), \theta'_r)$, $V \in \mathbf{V}$, converge to the same true joint observational distribution $\prod_{V\in\mathbf{V}}P(V|Pa(V),\theta)$ as for the correctly specified model $\mathcal{M}$. 
Therefore, the posterior predictive distribution converges to the true distribution, and its expected value is consistent. \qedsymbol
\end{proof}

\medskip \noindent This result is useful in practical applications, as choosing an instance from the right set of LVMs is less challenging than choosing  the exactly right LVM.  Therefore, given several candidate LVMs projecting on the same ADMG, we can rely on Occam's Razor  \cite{balasubramanian1997statistical,rasmussen2001occam} and favor the LVM with the simplest DAG structure.
%\end{methods}
%%%%%%%%%%%%%%%%%%%%%%%%%%%%%%%%%%%%%%%%%%%%%%%%
\begin{figure}[t]
\begin{center}
\begin{tabular}{c}
\includegraphics[scale=0.25]{img/legend.pdf}
\end{tabular}\\
\begin{tabular}{cc}
\includegraphics[scale=0.20]{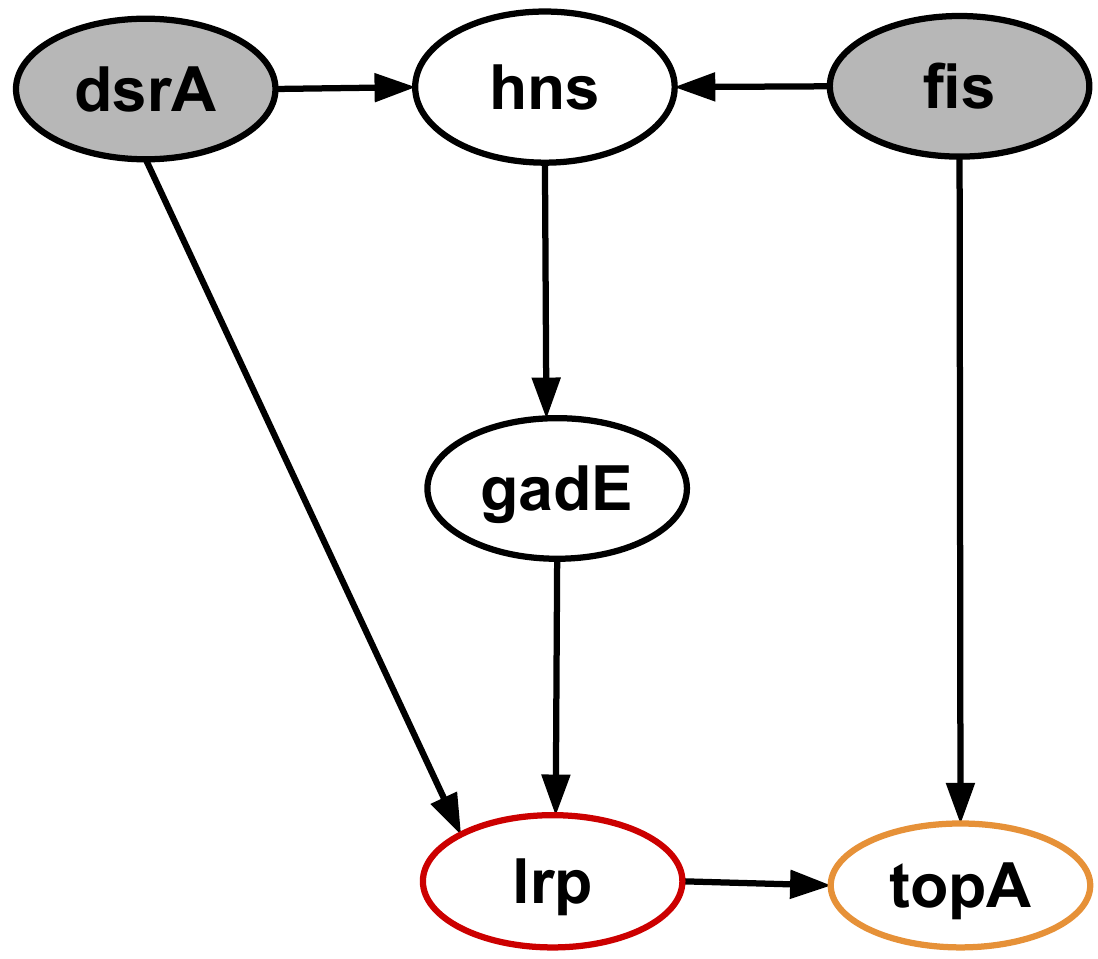} & 
\includegraphics[scale=0.20]{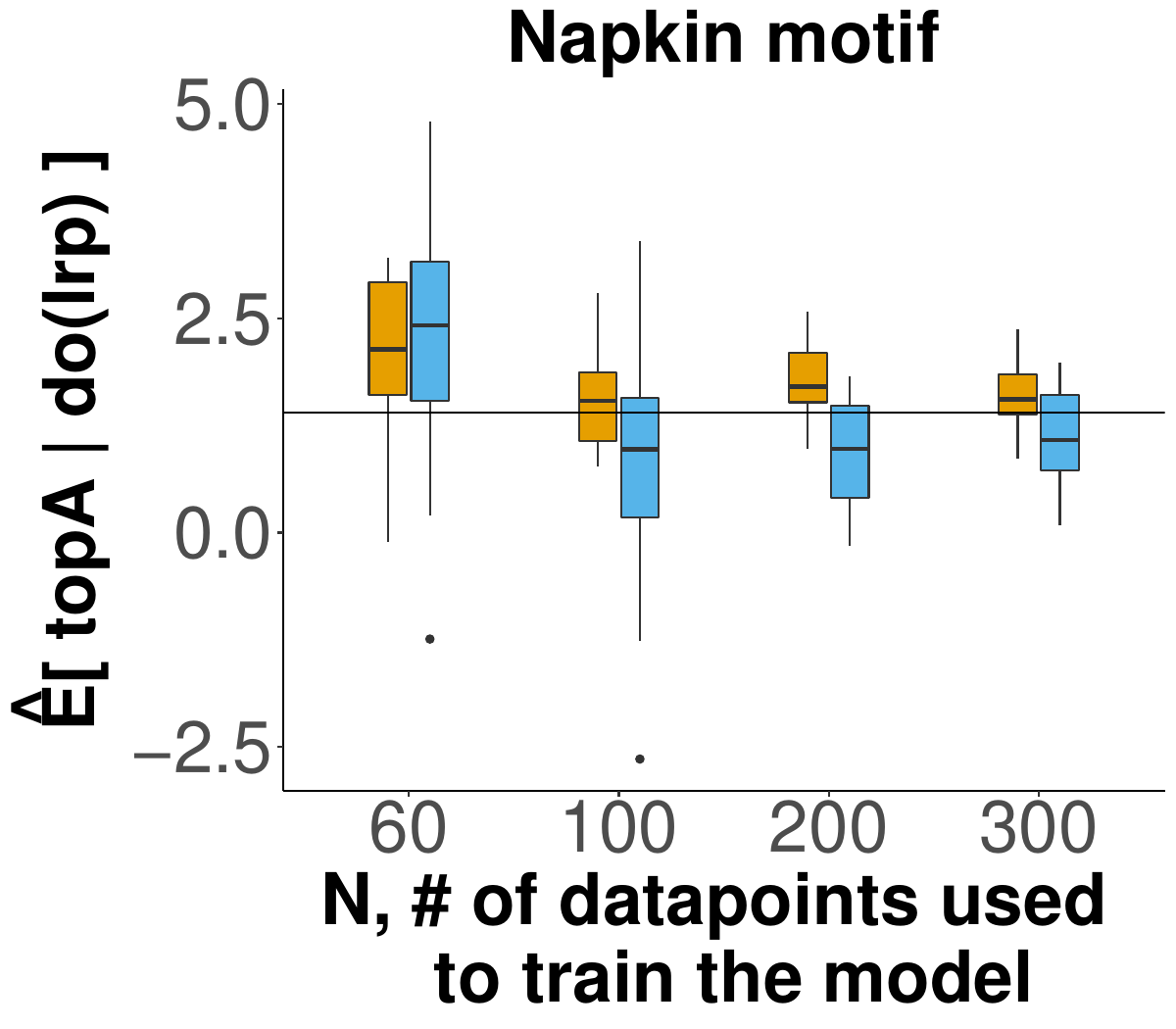} \\
(a) Case study 2 : Napkin motif & (b) $\hat{E}[topA| do(lrp = 0)]$
\end{tabular}
\end{center}
    \caption{\small 
     \textbf{Synthetic case study 2.} DAG labeled as in~\figref{fig:case1}. 
    (a) The Napkin network motif. 
    (b) Sampling distribution of $\hat{Q}_{\mathbf{x}}=\hat{E}[topA| do(lrp = 0)]$ over 20 observational datasets.
     \label{fig:case2}}
\end{figure}
%%%%%%%%%%%%%%%%%%%%%%%%%%%%%%%%%%%%%%%%%%%%%%
%%%%%%%%%%%%%%%%%%%%%%%%%%%%%%%%%%%%%%%%%%%%
%%%%%%%%%%%%%%%%%%%%%%%%%%%%%%%%%%%%%%%%%%%%%%%%%%%%%%%%%%%%%%%%%
%%%%%%%%%%%%%%%%%%%%%%%%%%%%%%%%%%%%%%%%%%%%%%%%%%%%%%%%%%%%%%%%%
%%%%%%%%%%%%%%%%%%%%%%%%%%%%%%%%%%%%%%%%%%%%%%%%%%%%%%%%%%%%%%%%%
\section{Case studies}
\subsection{Overview}
\label{strategy}

We illustrated the breadth and the practical utility of the proposed LVM-based estimation of identifiable causal queries in four synthetic and two experimental studies of biomolecular pathways, with topologies that challenge the existing methods for causal query estimation. 
We considered both the LVMs with the correct topology, and the LVMs with the correct topology for the observed variables but with a misspecified number of the latent variables.
Posterior distributions of the parameters were inferred with HMC in \texttt{Stan} \citep{stan2018rstan}.

The synthetic case studies illustrated the consistency of the causal queries of the form $Q_{\mathbf{X}}  = E_{\mathcal{M}_{\bar{X}}}[\mathbf{Y}|do(\mathbf{X} = \mathbf{x'})]$ 
(in the following we omit the subscript $\mathcal{M}_{\bar{X}}$, and state the value of $\mathbf{x'}$). The case studies incorporated a mix of probability distributions and a mix of informative and non-informative priors. 
Case studies 1,2, and 4 simulated observational data from parametric distributions with randomly selected values of parameters $\theta$.
Case study 3 generated data using stochastic differential equations.
The interventional data sets were obtained by sampling from the distribution with the true $\theta$ and the fixed targets of the interventions.  
To evaluate the performance of the proposed approach, the true values of $Q_{\mathbf{X}}$ were obtained by averaging 10,000 samples from the interventional data sets. 

The experimental case studies illustrated the accuracy of the causal queries of a different form, namely $Q_{\mathbf{X}}  = P_{\mathcal{M}_{\bar{X}}}(\mathbf{Y}|do(\mathbf{X} = \mathbf{x'}))$.
The experimental data were downloaded from Precision RNA-seq Expression Compendium for Independent Signal Exploration (PRECISE, \cite{sastry2019escherichia}). They contained 278 RNA-seq normalized expression profiles of {\it E. Coli} K-12 MG1655 and BW25113 across 154 unique experimental conditions. 
This manuscript focuses on pathways for which both observational and interventional data were available.
To evaluate the performance of the proposed approach, experimentally observed instances from $P(\mathbf{Y}|do(\mathbf{X} = \mathbf{x'}))$ were plotted against the estimated distributions.

Each case study was run on a single standard virtual machine on Google Cloud Platform with 2 vCPUs and 8 GB memory. Several virtual machine instances were used to run the case studies in parallel. The case studies took between 1.5 minutes and 1.8 hours.
%on a Google Cloud Platform.

%%%%%%%%%%%%%%%%%%%%%%%%%%%%%%%%%%%%%%%%%%%%%%%%%%%%%%%%%%%%%%%%%
\subsection{Synthetic case study 1:  The multi-cause feed-forward transcriptional regulatory network motif} 
\label{caseMulti}
\noindent \textbf{The system}  in \figref{fig:case1}(a) is an example of a common feed-forward network motif in \textit{E. coli} and many other prokaryotes~\citep{alon2019introduction}. 
The network was obtained by querying the EcoCyc database~\citep{keseler2021ecocyc} to discover which  front door motifs with one or more confounders  and one or more causes exist in \textit{E. coli}. More than 10,000 different cases were found. For this case study we randomly selected one case. Despite being ubiquitous, the case study is challenging because it has multiple causes.

\noindent \textbf{Query of interest}
$Q_{\mathbf{X}} = E[cas2|do(\mathbf{X} = 0)]$, where $\mathbf{X} = \{dsrA, gadX, fis\}$. In this query the back-door criterion does not hold but the front-door criterion holds. 

\noindent \textbf{Data} of the latent variables followed a Normal distribution, and the remaining variables a Bernoulli distribution with logit parameterization.

\noindent \textbf{LVM with correct topology} assumed the correct data generation process with non-informative $\mathcal{N}(0,10)$ priors over all the parameters. 

\noindent \textbf{LVM with misspecified number of latent variables} wrongly assumed only one latent variable.

%%%%%%%%%%%%%%%%%%%%%%%%%%%%%%%%%%%%%%%%%%%%%%%%%%%%%%%%%%%%%%%%%
\subsection{Synthetic case study 2 :  The Napkin motif}
\label{case4}
\textbf{The system} 
in \figref{fig:case2}(b) is called the second Napkin problem in \citep{pearl2018book}.
The network was obtained by querying the EcoCyc database~\citep{keseler2021ecocyc} to discover all napkin motifs with two or more confounders in \textit{E. coli}. More than 10,000 different cases were found. For this case study we randomly selected one case.

\noindent \textbf{Causal query of interest}
$Q_{lrp} = E[topA|do(lrp=1)]$.
The system requires a non-trivial application of the do-calculus, because we cannot block the back-door path from $lrp$ to $topA$ ($hns$ is a collider and $gadE$ is an ancestor of a collider), and because the front-door criterion does not hold (there is no mediator between $lrp$ and $topA$)~\citep{helske2020estimation, hughes1998cd4, pearl2018book, jung2020learning}.

\noindent \textbf{Data} of $hns$ was modeled with a gamma distribution (representative of expression measurements with a fluorescent reporter). The expression of all the other genes $dsr$, $fis$, $gadE$ and $topA$ were modeled with Gaussian distributions (representative of measurements or relative expression, such as with RT-PCR).
$lrp$ followed a Bernoulli distribution with logit parametrization.

\noindent \textbf{LVM with correct topology} assumed the correct data generation process with non-informative $\mathcal{N}(0,10)$ priors over all the parameters.

\noindent \textbf{LVM with misspecified number of latent variables} wrongly assumed two latent variables between $hns$ and $topA$.

%%%%%%%%%%%%%%%%%%%%%%%%%%%%%%%%%%%%%%%%%%%%%%%%%%%%%%%%%%%%%%%%%%%%%%%

%%%%%%%%%%%%%%%%%%%%%%%%%%%%%%%%%%%%%%%%%%%%%%%%
%\vspace{-0.3cm}
\subsection{Synthetic case study 3: The signaling model}
\label{caseStudy1}
\begin{figure}[t]
\begin{center}
\begin{tabular}{c}
\includegraphics[scale=0.25]{img/legend.pdf}
\end{tabular}\\
\begin{tabular}{cc}
\includegraphics[scale=0.2]{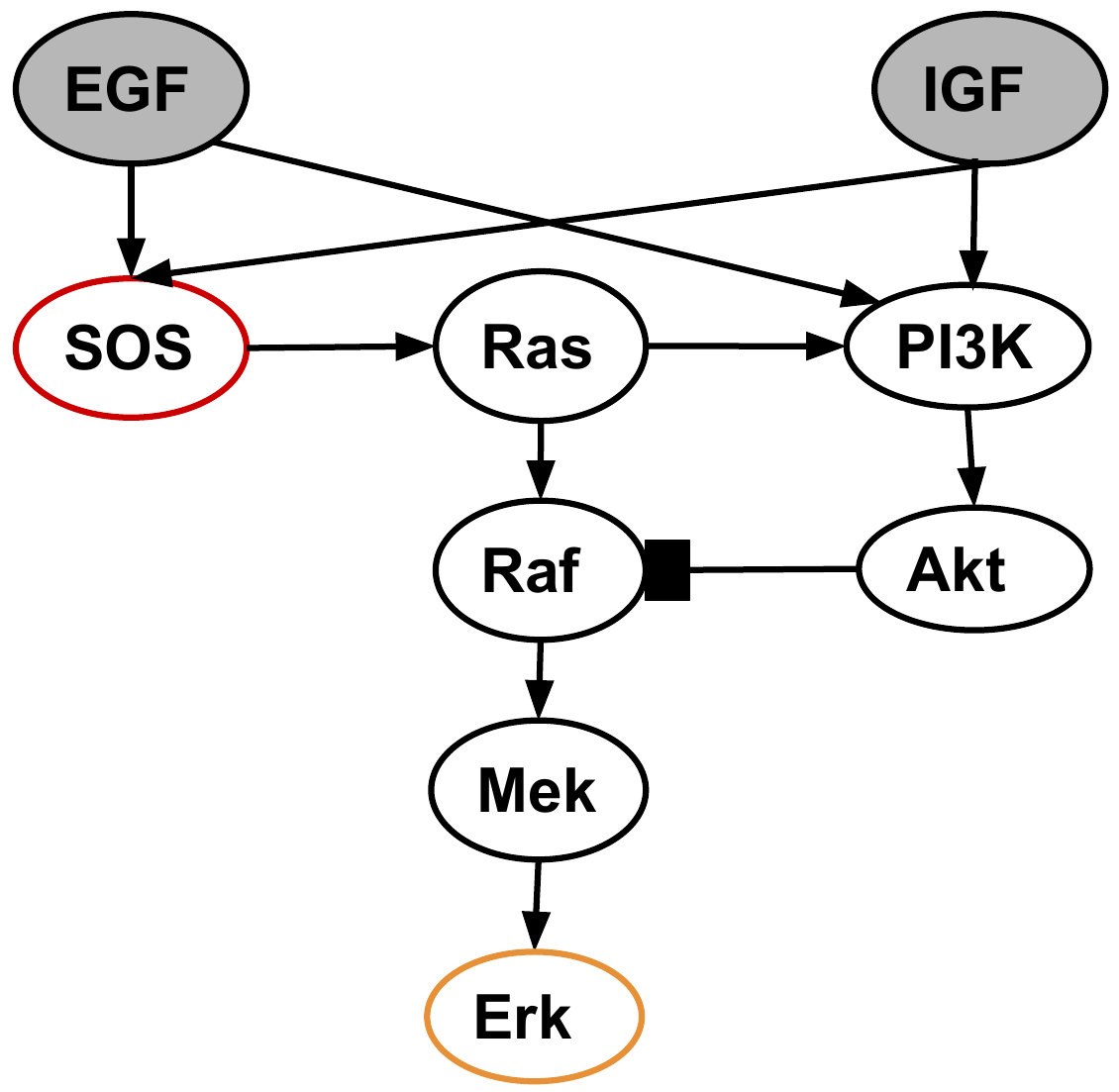}&
\includegraphics[scale=0.20]{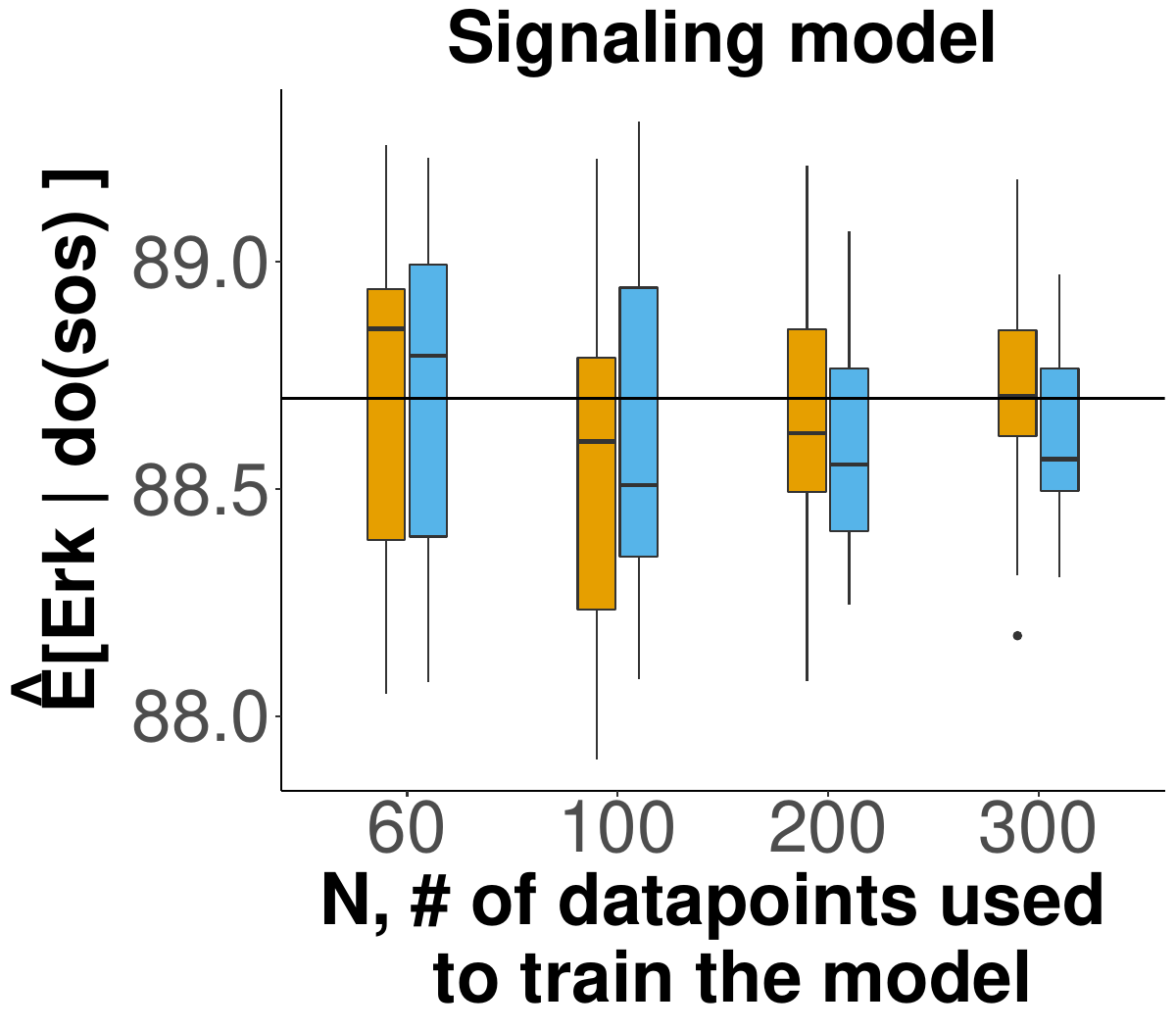}\\
(a) Case study 3 : Signaling model & (b) $\hat{E}[Erk| do(SOS = 70)]$
\end{tabular}
\end{center}
    \caption{\small 
     \textbf{Synthetic case study 3.} DAG labeled as in~\figref{fig:case1}. 
    (a) The signaling model. Nodes are proteins, pointed/flat-headed edges are relationships of type \textit{increase}/\textit{decrease}.
    (b) Sampling distribution of $\hat{Q}_{\mathbf{x}}=\hat{E}[Erk| do(SOS = 70)]$ over 20 observational datasets.
     \label{fig:case3}}
\end{figure}

{\bf The system} in \figref{fig:case3}(a) is a well-studied insulin-like growth factor signaling system regulating growth and energy metabolism of a cell~\citep{9328353}.
IGF and EGF are latent. 

\noindent \textbf{Causal query of interest}
$Q_{{SOS}} = E[Erk|do(SOS=70)]$. Similarly to case study 2, $Q_{{SOS}}$ does not satisfy the back-door or the front-door criteria.  

\noindent \textbf{Data} mimicked the experimental process of collecting observational and interventional data. Since dynamics of this system are well characterized in form of stochastic differential equations (SDE) \citep{bianconi2012computational}, we generated observational
data by simulating from the SDE. We set the initial amount of each protein molecule to 100, and generated subsequent observations via the Gillespie algorithm~\citep{gillespie1977exact} in the \textit{smfsb}~\citep{wilkinson2018package} R package. Replicates were generated by randomly initializing EGF and IGF.
Interventional data were generated similarly, while fixing SOS=70. 

\noindent \textbf{LVM with correct topology} Unlike in the previous case studies, the variables were not modeled following the data generation process, but only approximated it.
The exogenous variables were modeled with a Gaussian distribution.
The rest of the variables were modeled by representing the biomolecular reactions with a Hill function, as common in the biological practice \noindent \citep{alon2019introduction}, and were approximated with a sigmoid function as follows,
$\mathcal{N}(\frac{100}{1 + exp(\mathbf{\theta}^{T}  Pa(X) + \theta_0)}, \sigma_X)$.
For a node $X$ with $q$ parents, $Pa(X)$ was a $q \times 1$ vector of measurements on the parent nodes, $\mathbf{\theta}^{T}$ was a $1 \times q$ vector of unknown parameters, and $\theta_0$ was an unknown scalar parameter. 
The non-informative $\mathcal{N}(0,10)$ priors of the parameters $\theta$ in the sigmoid had a constraint of being positive for the relationships of type increase and negative for relationships of type decrease.

\noindent \textbf{LVM with misspecified number of latent variables}
only included EGF as latent, and omitted IGF.

%%%%%%%%%%%%%%%%%%%%%%%%%%%%%%%%%%%%%%%%%%%%%%%%
%%%%%%%%%%%%%%%%%%%%%%%%%%%%%%%%%%%%%%%%%%
%%%%%%%%%%%%%%%%%%%%%%%%%%%%%%%%%%%%%%%%%%%%%%%%%%%%%%%%%%%%%%%%%

\subsection{Synthetic case study 4: The SARS-CoV-2 model}
\label{case3}

\begin{center}
\begin{figure*}[t]
\begin{center}
\begin{tabular}{cc}
$~~~~~~~~~~~~~~~~~~~~~~~~~~~~~~~~~~~~~~~~~~~~~~~~~~~~~~~~~~~~~~~~~~~~~~~~~~~~~~~~~~~~~~~~~$ &
\includegraphics[scale=0.25]{img/legend.pdf}
\end{tabular}
\end{center}
\begin{tabular}{ccc}
\includegraphics[scale=0.18]{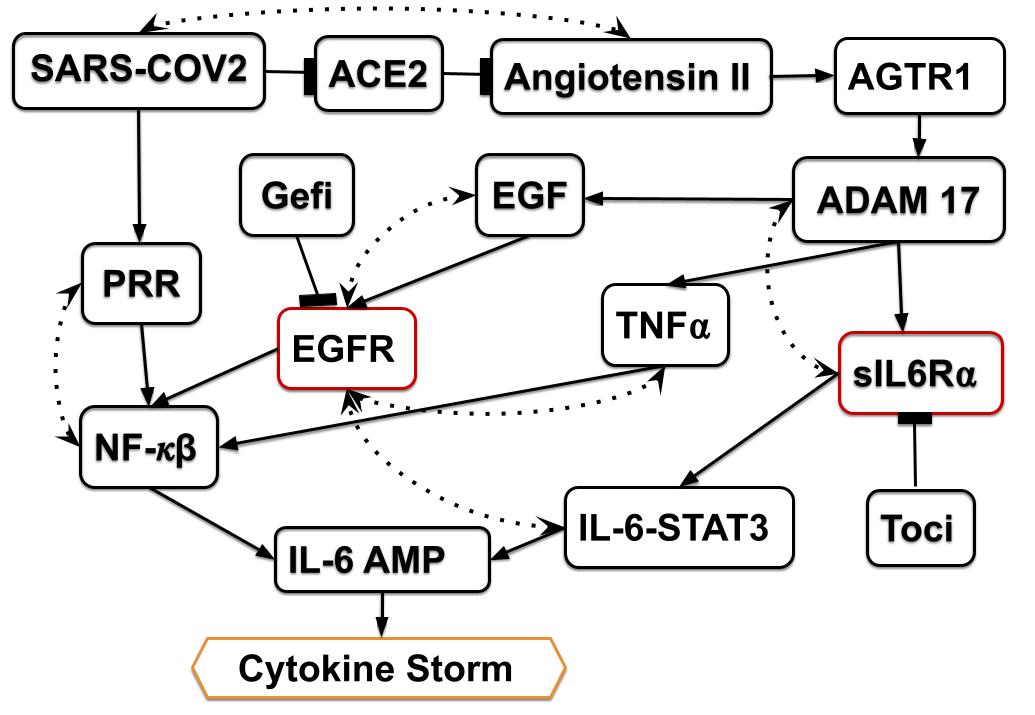} &
\includegraphics[scale=0.25]{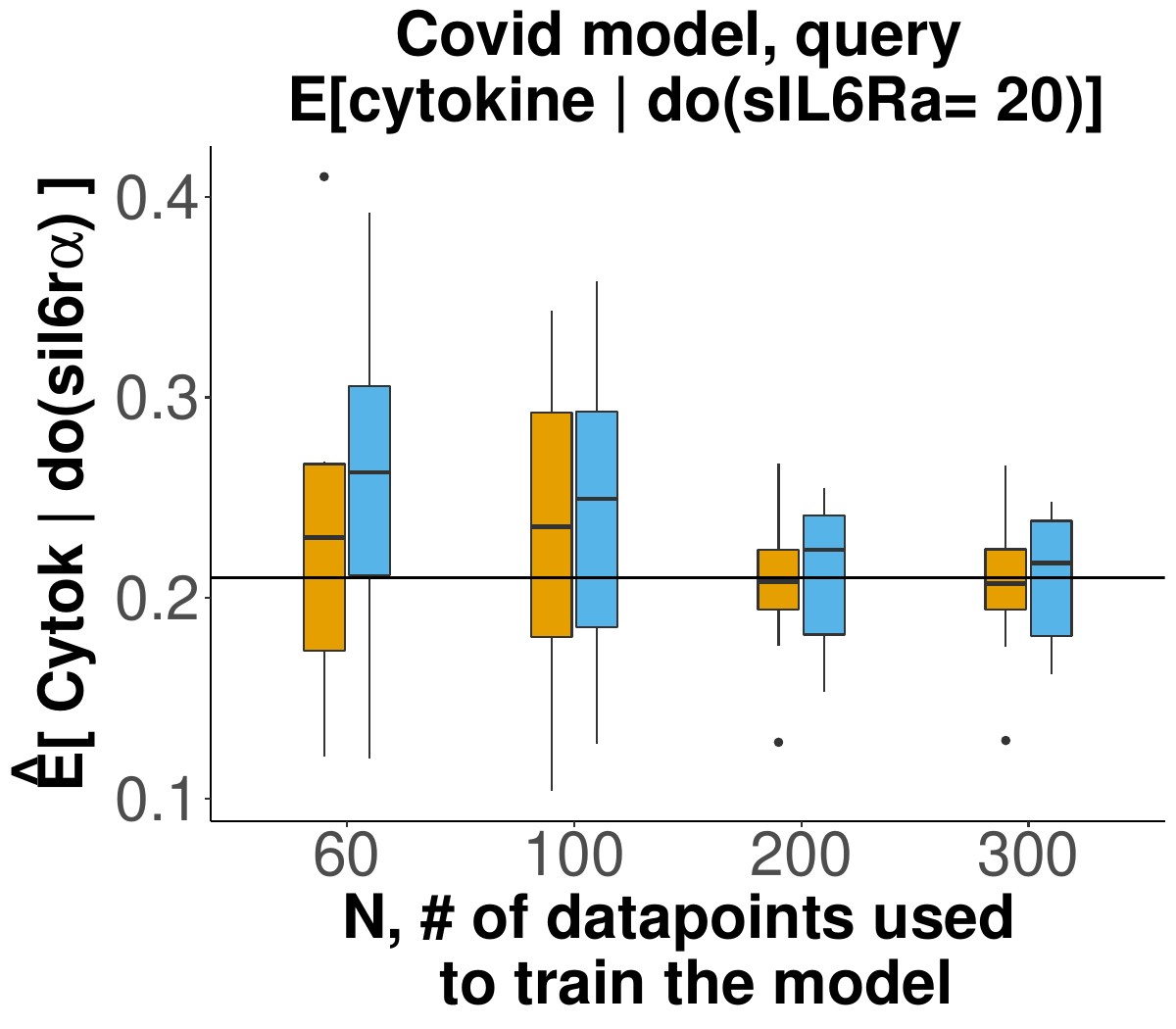} &
\includegraphics[scale=0.25]{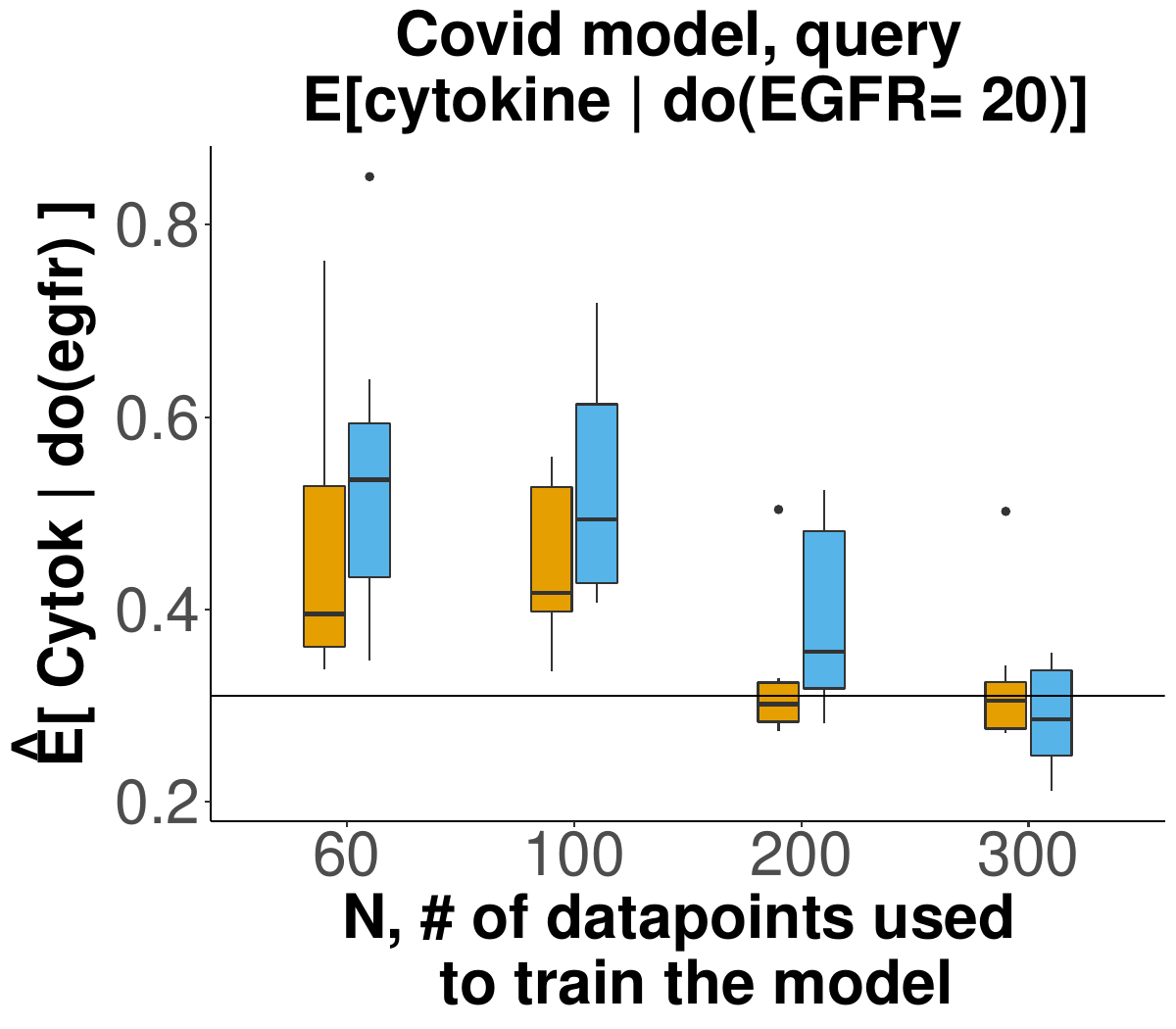} 
\\
(a) Case study 4 : Covid model & (b) & (c)
\end{tabular}
    \caption{\small   
    \textbf{Synthetic case study 4.} DAG labeled as in~\figref{fig:case1}.
    (a) The SARS-CoV-2 model. Dotted edges indicate presence of latent variables. sIL6R$\alpha$ and EGFR are targets of intervention, Cytokine Storm is the effect. 
    (b) Sampling distribution of $\hat{Q}_{\mathbf{x}}=\hat{E}[Cytokine | do(sIL6R\alpha= 20)]$ over 20 observational datasets. 
    (c) As in (b), for $\hat{Q}_{\mathbf{x}}=\hat{E}[Cytokine | do(EGFR= 20)]$. 
    \label{fig:case4_covid}}
\end{figure*}
\end{center}

%\textbf{The system} 
\noindent \textbf{The system} in \figref{fig:case4_covid}(a) models activation of Cytokine Release Syndrome
(Cytokine Storm), known to cause tissue damage in severely ill SARS-CoV-2 patients \citep{ulhaq_2020}. 
The simultaneous activation of the NF-$\kappa$B and IL6-STAT3 activates IL6-AMP, which in turn activates Cytokine Storm \citep{hirano_2020}. 
The system showcases the ability of a causal LVM to estimate multiple causal queries after a single instance of training.

The network was extracted from COVID-19 Open Research Dataset (CORD-19) document corpus using the Integrated Dynamical
Reasoner and Assembler (INDRA)  workflow \citep{9328353}, and by quering and expressing the corresponding causal statements in the Biological Expression Language (BEL) using PyBEL~\cite{hoyt2017pybel}. Presence of latent variables was determined by querying pairs of entities in the network for common causes in the corpus.

\noindent \textbf{Causal queries of interest} examine the ability of two different drugs to prevent Cytokine Storm.
Tocilizumab (Toci) is an immunosuppressive drug that targets sIL6R$\alpha$ and blocks the IL6 signal transduction
pathwa \citep{zhang_2020}.
The first causal query examined the effect of Toci by setting its target sIL6R$\alpha$=20 (low value), i.e. $Q_{{sIL6R\alpha}}=E[Cytokine | do(sIL6R\alpha)= 20)]$. The query is identifiable using the backdoor criterion. The drug Gefitinib (Gefi) blocks $EGFR$. The second causal query examined the effect of Gefi, i.e. $Q_{{EGFR}}=E[Cytokine | do(EGFR)= 20)]$.  The query is not identifiable via either the backdoor or the front-door criterion, but is identified via the do-calculus. 

\noindent \textbf{Data} of the latent variables had Gaussian distributions, Cytokine storm had a Bernoulli distribution with logit parameterization, and the remaining variables were simulated with a Hill function as in case study 3. 

\noindent \textbf{LVM with the correct topology} assumed the correct data generation process where it contained two latent variables between 
(SARS-CoV-2
and 
Angiotensin II),
(ADAM17
and 
sIL6R$\alpha$),
and (PRR and 
NF-$\kappa$B),
and one latent variable for each remaining dotted edge. A mixture of Non-informative $\mathcal{N}(0,10)$ and informative priors $\mathcal{N}(E[\theta],1)$ where $E[\theta]$ were between 20-45 was used.

\noindent \textbf{LVM with misspecified number of latent variables } wrongly assumed only one latent variable for each dotted edge.
%%%%%%%%%%%%%%%%%%%%%%%%%%%%%%%%%%%%%%%%%%%%%%%%
%%%%%%%%%%%%%%%%%%%%%%%%%%%%%%%%%%%%%%%%%%%%%%%%%%%%

\subsection{Experimental case study 5: The single-cause feed-forward transcriptional regulatory network motif}
\label{case5}
\begin{figure}[t]
\begin{center}
\begin{tabular}{c}
\includegraphics[scale=0.3]{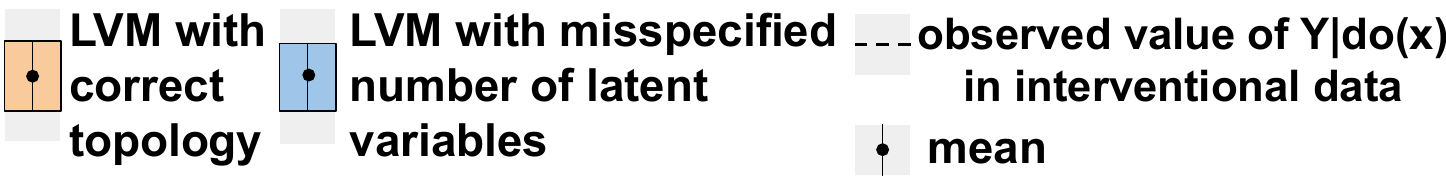}
\end{tabular}
\begin{tabular}{cc}
\includegraphics[scale=0.2]{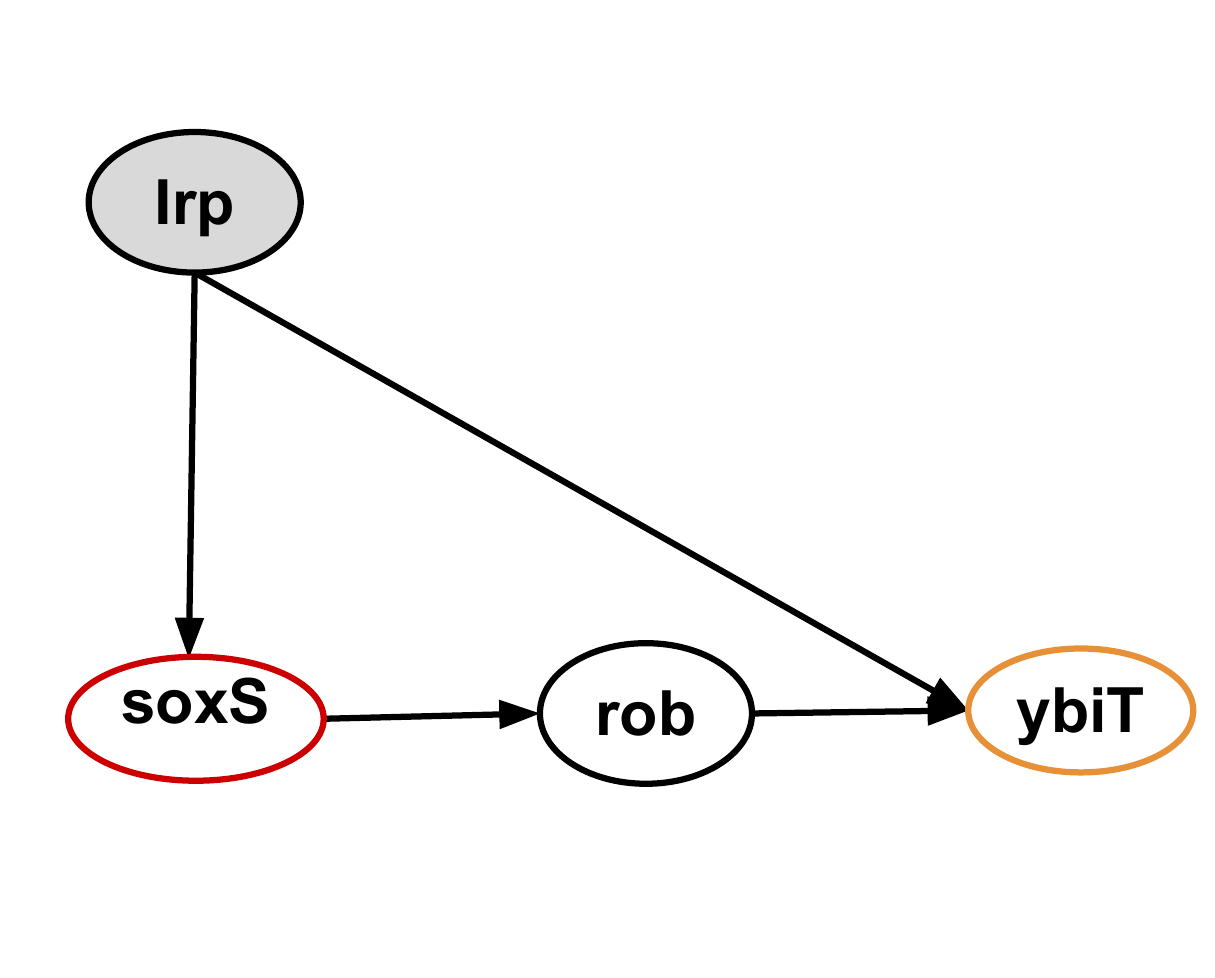}&
\includegraphics[scale=0.2]{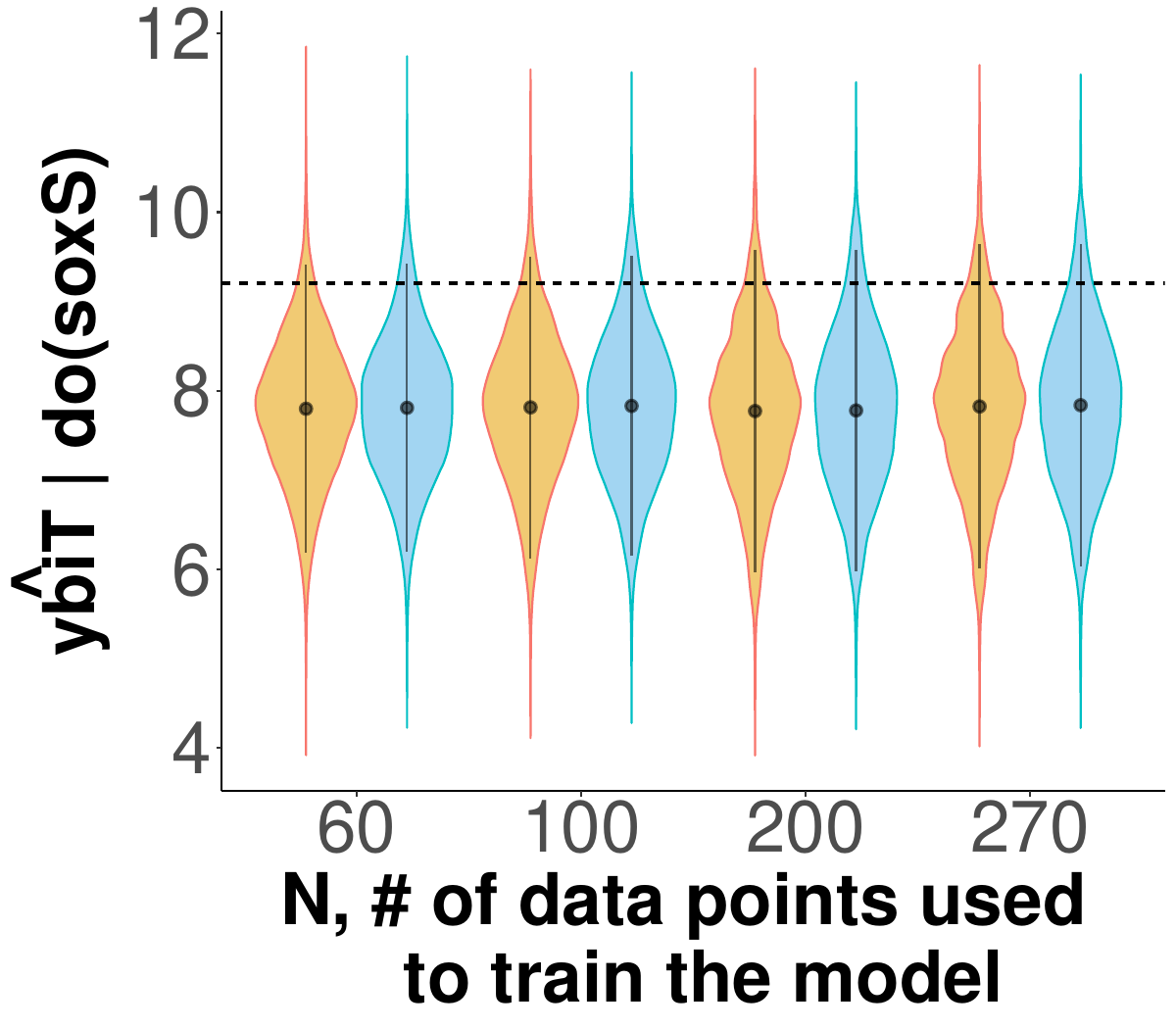}\\
(a) & (b)
\end{tabular}
\end{center}
    \caption{\small 
     \textbf{Experimental case study 5.} (a) The transcriptional regulatory network with the single-cause feed-forward motif. (b) The causal query in form of a probability distribution $\hat{Q}_{soxS} = \hat{P}(ybiT | do(soxS)=0)$. 
     %The distributions accurately represent the observed interventional data. 
     The two (overlapping) horizontal dashed lines indicate two observed values of $ybiT|soxS=0$ in the interventional experiments.
     \label{fig:case5}}
\end{figure}

\textbf{The system} in \figref{fig:case5}(a) is a common feed-forward network motif in the transcriptional regulatory network of {\it E. Coli}, where the effect variable is not a direct effect of the cause variable.
The network was obtained  by querying  the EcoCyc database \cite{keseler2021ecocyc} with the Pathway-tools lisp api \cite{karp2021pathway} for all 3-hop ancestors of all 2-hop descendants of the genes with available experimental interventional data. 5800 such cases were found. We randomly selected the pathway in \figref{fig:case5}(a).
This system is similar to case study 1, but with a single cause and a single latent variable. 

\noindent \textbf{Causal query of interest} $Q_{soxS}=P(ybiT | do(soxS=0))$. Although $lrp$ is latent, the mediator $rob$  made the query identifiable according to the front-door criterion.

\noindent \textbf{Experimental data} contained 278 RNA-seq normalized expression profiles of {\it E. Coli} K-12 MG1655 and BW25113 across 154 unique experimental conditions. 
Interventional data corresponded to the query of interest, i.e. $soxS = 0$. We used this data to evaluate the performance of the proposed approach. The observational and interventional data was obtained from the PRECISE database~\citep{sastry2019escherichia}.

\noindent \textbf{LVM with correct topology} specified Gaussian distributions with non-informative priors $N(0,10)$.

\noindent \textbf{LVM with misspecified number of latent variables} wrongly assumed two latent variables.
%%%%%%%%%%%%%%%%%%%%%%%%%%%%%%%%%%%%%%%%%%%%%%%%%%%%%%%%%%%%%%%%%
%%%%%%%%%%%%%%%%%%%%%%%%%%%%%%%%%%%%%%%%%%%%%%%%%%%%%%%%%%%%%%%%%
\subsection{Experimental case study 6: The Napkin motif}
\label{case6}

\begin{figure}[t]
\begin{center}
\begin{tabular}{c}
\includegraphics[scale=0.3]{img/legendCase5.pdf}
\end{tabular}
\begin{tabular}{cc}
\includegraphics[scale=0.2]{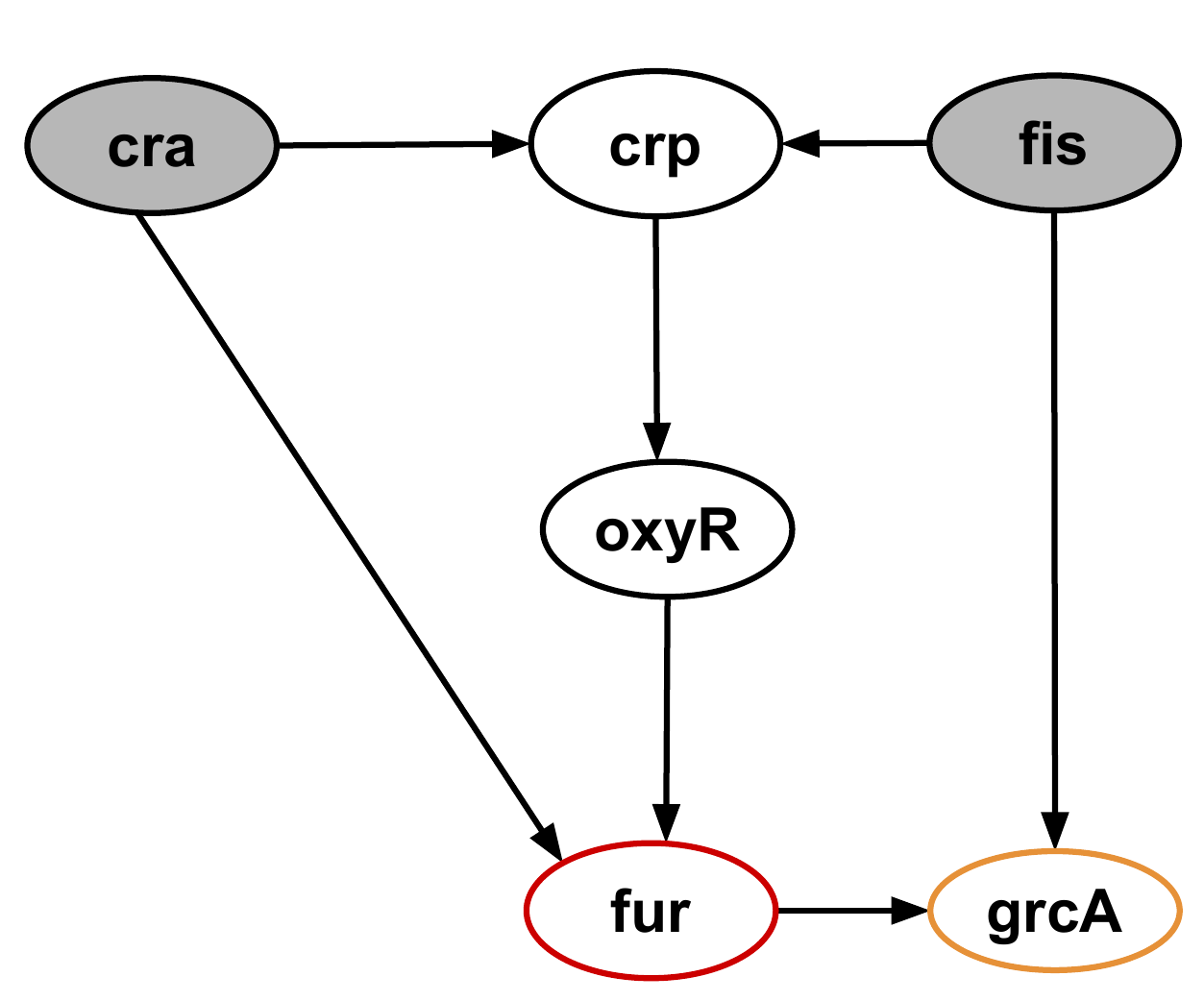}&
\includegraphics[scale=0.2]{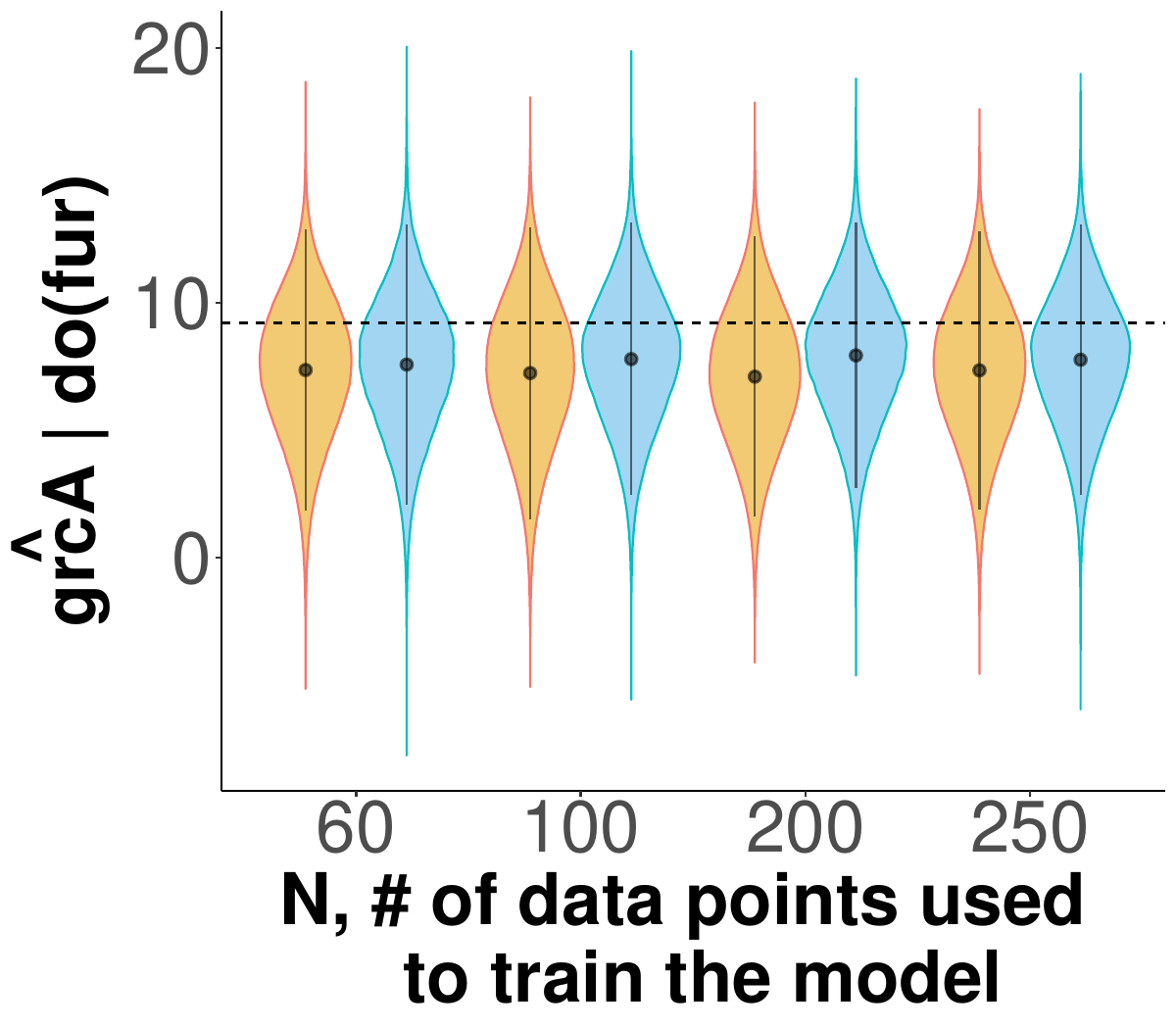}\\
(a) & (b)
\end{tabular}
\end{center}
    \caption{\small 
     \textbf{Experimental case study 6.} (a) The transcriptional regulatory network with the Napkin motif. (b) The causal query in form of a probability distribution  $\hat{Q}_{fur} = \hat{P}(grcA | do(fur = 0))$.
     The two (overlapping) horizontal dashed lines indicate two observed values of $grcA|fur=0$ in the interventional experiments.
     \label{fig:case6}}
\end{figure}

\noindent \textbf{The system} in \figref{case6} is the same system as in case study 2.
The network was obtained  by querying  the EcoCyc database \cite{keseler2021ecocyc} with the Pathway-tools lisp api \cite{karp2021pathway} for all 3-hop ancestors of all 2-hop descendants of the genes with available experimental interventional data. 5500 such cases were found. We randomly selected the pathway in \figref{fig:case6}(a).

\noindent \textbf{Causal query of interest} $Q_{fur} = P(grcA|do(fur=0))$.

\noindent \textbf{Experimental data} were as in case study 5. 
Interventional data corresponded to the query of interest, i.e. $fur = 0$. We used this data to evaluate the performance of the proposed approach.

\noindent \textbf{LVM with correct topology} assumed a Gaussian distribution over all the variables with non-informative priors $N(0,10)$.

\noindent \textbf{LVM with misspecified number of latent variables} wrongly assumed two latent variables between $crp$ and $grcA$.

%%%%%%%%%%%%%%%%%%%%%%%%%%%%%%%%%%%%%%%%%%%%%%%%%%%%%%%%%%%%%%%%%
\section{Results}

\textbf{In the synthetic case studies with correct LVM topologies, the estimates $\hat{E}[Y|do(X = x')]$ were consistent}
\figref{fig:case1}(b), \figref{fig:case2}(b), \figref{fig:case3}(b), and \figref{fig:case4_covid}(b,c) show sampling distributions of 20 $\hat{E}[Y|do(X = x')]$, summarized from 20 repetitions of generating observational data with $N$ replicates and estimating the causal query (orange boxes). 
Although the expected values and the variances of the sampling distributions depended on the data and on the system, all the estimates approached the true value with reduced variability as $N$ increased.
This was the case despite the diverse topologies of the networks, and despite the diverse distributional assumptions.

\medskip \noindent \textbf{In the synthetic case studies with LVM with misspecified number of latent variables, the estimates $\hat{E}[Y|do(X = x')]$ were consistent}
\figref{fig:case1}(b), \figref{fig:case2}(b), \figref{fig:case3}(b), and \figref{fig:case4_covid}(b,c) show sampling distributions of 20 $\hat{E}[Y|do(X = x')]$, summarized from 20 repetitions of generating observational data and estimating the causal query with LVMs specifying a wrong number of latent variables (blue boxes). 
While these sampling distributions had more bias and variance than the distributions from the correctly specified LVMs for small $N$, they approached the true values with reduced variation as $N$ increased. 

\medskip \noindent \textbf{In the experimental case studies, the estimates $\hat{P}(Y|do(X = x'))$ accurately represented the observed interventional data}
\figref{fig:case5}(b) and \figref{fig:case6} (b) display the estimated query specified in a different form, namely  posterior interventional distributions $\hat{P}(Y|do(X = x'))$, as function of the number of observations used to train the LVM. The horizontal lines correspond to two experimental interventional measurements. As the values were very similar, the lines overlap. Despite real-life experimental artifacts, such as dynamic range compression and measurement errors that affected the observational data, and despite the approximate nature of the modeling assumptions, the estimated distributions covered the values observed from the experimental interventional data.

\medskip \noindent \textbf{In both synthetic and experimental case studies, the estimates were accurate despite the approximate nature of the parametric assumptions} A common criticism of LVMs is the requirement of parametric distributional assumptions. Despite the criticism, in case study 3 the estimate  $\hat{E}[Y|do(X = x')]$ was  consistent, and in case studies 5 and 6 the estimates $\hat{P}(Y|do(X = x'))$ were accurate, even though the LVMs  imperfectly approximated the unknown data generating distributions.

\medskip \noindent \textbf{The proposed approach expanded the current use of LVMs for estimating causal queries} 
All the case studies showed the accuracy
of LVM-based estimation in situations where LVM-based estimators
have so far not been traditionally applied, i.e. pathways without proxy
variables ~\citep{louizos2017causal, kuroki2014measurement}, or with multiple causes~\citep{ wang2019blessings}.

\medskip \noindent \textbf{Non-LVM-based approaches could not be applied to any case study in this manuscript.} An alternative to LVM-based estimators are non-parametric or semi-parametric estimators. Unfortunately, we could not apply any of these methods to the case studies in this manuscript, as their implementations were either not publicly available\cite{jung2020learning} or could not handle continuous or multi-cause queries\cite{bhattacharya2020semiparametric}.
However, in biomolecular systems multiple causes and effects are common, and the variables are not always discrete. All the case studies in this manuscript had either multiple causes or a continuous cause. They demonstrated the utility and the accuracy of LVM-based estimation in these situations.

\medskip \noindent \textbf{Synthetic case studies 1, 2 and experimental case studies 5, and 6 represented commonly occurring patterns in biomolecular pathways} To demonstrate the ubiquity of the  feed-forward motif presented in case studies 1 and 5, we queried the EcoCyc database~\citep{keseler2021ecocyc} for all~\textit{E. coli} front-door motifs with one or more confounders and one or more causes, and found over 1000 such motifs. To demonstrate the ubiquity of the motif in case studies 2 and 6, we queried the EcoCyc database for all~\textit{E. coli} Napkin motifs with two or more confounders, and found over 1000 such motifs.

To further illustrate the ubiquity of network motifs that satisfy the front-door criterion of studies 1 and 5 in other organisms, we queried the repository INDRA \cite{gyori2017word} for all such motifs in humans. We applied strict quality filters to ensure each causal relationship was supported by at least 5 publications. We found 90 such instances.  Reference \cite{mangan2003structure} provides additional examples of the feed-forward motif in yeast. Overall, the case studies illustrate the ability of the proposed approach to quantitatively answer potentially large numbers of important biological questions, limited only by qualitative prior knowledge of the organism's regulatory network and the availability of experimental data.

\medskip \noindent \textbf{In synthetic case study 3, LVM-based estimators estimated multiple queries from a single trained model}
After training the LVM once, the proposed approach enabled the estimation of two distinct causal queries.
This is particularly valuable for probabilistic reasoning systems, e.g. in systems biology or medical diagnosis, where large-scale models are expensive to train and maintain. 
This is not possible with non-LVM-based estimators, which require us to derive a new statistical estimand for each causal query anew.

\section{Discussion}

A major criticism of traditional pathway modeling is its inability to account for external influences on pathway components.  
This is particularly relevant to causal inference, as ignoring the effect of unobserved confounding can undermine the inaccuracy of the results.  
In this manuscript, we advocate for the explicit use of LVMs, and for applying Pearl's do-calculus to determine whether the causal effect can be identified.

We proposed training the LVMs with exact inference algorithms as they guarantee asymptotically exact samples. These algorithms are computationally expensive. However, trained on observational data once, an LVM can estimate multiple causal queries corresponding to multiple mutilated versions of the original DAG. After training, the estimation of causal query is instantaneous. This indicates that with enough experimental replicates and computational resources the proposed approach can be scaled to larger networks.

We showed that LVM-based estimation of identifiable causal queries is successful in situations that challenge other statistical estimators, e.g. in presence of interventions on continuous variables, and queries with multiple causes and effects. 
The estimation is robust to latent variable misspecification, 
and
to parametric approximations of complex processes of data generation. 
As all these situations are very common, the proposed approach expands the feasibility and scope of causal inference in biomolecular pathways.

%%%%%%%%%%%%%%%%%%%%%%%%%%%%%%%%%%%%%%%%%%%%%%%%%%%%%%%%%%%%%%%%%
%%%%%%%%%%%%%%%%%%%%%%%%%%%%%%%%
%%%%%%%%%%%%%%%%%%%%%%%%%%%%%%%%%%%%%%%%%%%%%%%%%%%%%%%%%%%%%%%%%%%%%
%%%%%%%%%%%%%%%%%%%%%%%%%%%%%%%%%%%%%%%%%%%%%%%%%%%%%%%%%%%%%%%%%
\section*{Funding}
JZ is supported by the PNNL Directed R\&D-funded Data-Model Convergence Initiative. PNNL is operated for the DOE by Battelle Memorial Institute under Contract DE-AC05-76RLO1830.
CTH is supported by the DARPA Young Faculty Award W911NF2010255 (PI: Benjamin M. Gyori).
KS is supported by Muscular dystrophy association grant \# 574137 and Answer ALS consortium.
OV acknowledges the support of NSF-BIO/DBI 1759736, NSF-BIO/DBI 1950412, NIH-NLM-R01 1R01LM013115 and of the Chan-Zuckerberg foundation.\vspace*{-12pt}
%%%%%%%%%%%%%%%%%%%%%%%%%%%%%%%%%%%%%%%%%%%%%%%%%%%%%%%%%%%%%%%%%

\bibliographystyle{unsrtnat}
\bibliography{references}  %%% Uncomment this line and comment out the ``thebibliography'' section below to use the external .bib file (using bibtex) .

\clearpage
\appendix
{\Large \bf Appendix}

\end{document}